%% file: main.tex
\documentclass[10pt,twocolumn,letterpaper]{article}

\usepackage{cvpr}              % To produce the CAMERA-READY version
\input{preamble}

\definecolor{cvprblue}{rgb}{0.21,0.49,0.74}
\usepackage[pagebackref,breaklinks,colorlinks,allcolors=cvprblue]{hyperref}
\usepackage{url}

\def\paperID{8412} % *** Enter the Paper ID here
\def\confName{CVPR}
\def\confYear{2025}

\title{EIDT-V: Exploiting Intersections in Diffusion Trajectories for Model-Agnostic, Zero-Shot, Training-Free Text-to-Video Generation}

\makeatletter
\newcommand\blfootnote[1]{%
  \begingroup
  \renewcommand\thefootnote{}%
  \footnote{#1}%
  \addtocounter{footnote}{-1}%
  \endgroup
}
\makeatother

\author{
Diljeet Jagpal$^{1}$, Xi Chen$^{1,2,*}$, and Vinay P. Namboodiri$^{1}$\\[1ex]
$^{1}$University of Bath; $^{2}$Fudan University\\[1ex]
{\tt\small \{dkjj20, xc841, vpn22\}@bath.ac.uk, x\_chen@fudan.edu.cn}
}

\begin{document}
\twocolumn[{%
\renewcommand\twocolumn[1][]{#1}%
\maketitle
\vspace{-0.5cm}
\begin{center}
    \centering
    \input{Figures/main}
\end{center}%
}]

\blfootnote{* Corresponding author.}
\blfootnote{Supported by EPSRC (EP/T518013/1, EP/Y021614/1); AAM available under a CC BY licence.}

\input{sec/0_abstract}
\input{sec/_1_Introduction}
\input{sec/_2_RelatedWorks}

\input{sec/_3_Background}
\input{sec/_4_Methodology}
\input{sec/_5_Results}
\input{sec/_6_Discussion}
\input{sec/_7_Conclusions}

%\input{sec/1_intro}
%\input{sec/2_formatting}
%\input{sec/3_finalcopy}

%%%
\newpage

% paper for review cannot have any acknowledgement section
\input{sec/_8_Acknowledgments}

{
    \small
    \bibliographystyle{ieeenat_fullname}
    \bibliography{main}
}


\input{X_suppl}

\end{document}

%% file: preamble.tex
\usepackage{multirow}
\usepackage{float}
\usepackage{amssymb}
\newcommand{\cmark}{\checkmark} % Tick mark
\newcommand{\xmark}{\texttimes} % Cross mark

\usepackage{lineno}
\usepackage[accsupp]{axessibility}  % Improves PDF readability for those with disabilities.

%% file: Figures/main.tex
\centering
\includegraphics[width=\textwidth]{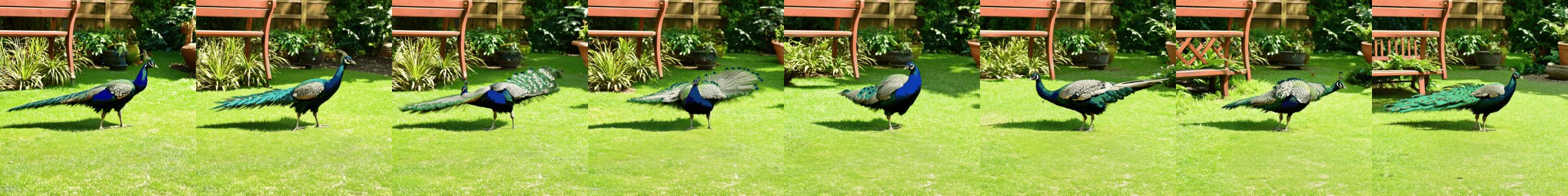}
\includegraphics[width=\textwidth]{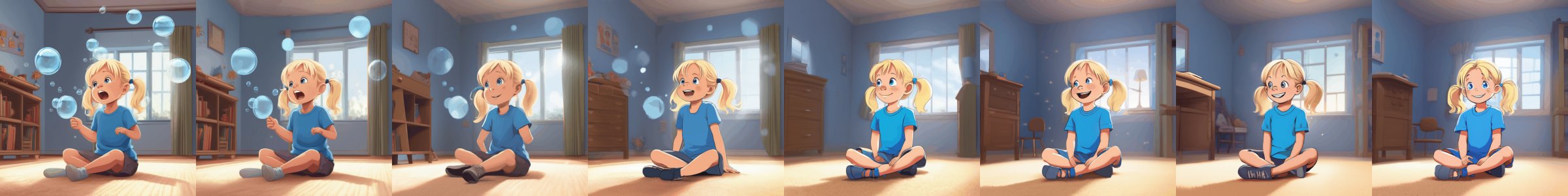}
\small
Eight equally spaced frames from 24-frame GIFs generated by our EIDT-V model. Top row shows SD3 Medium \cite{esser_scaling_2024} results for prompt: \textbf{``A peacock displaying its feathers''}. Bottom row shows SDXL \cite{podell_sdxl_2023} results for prompt: \textbf{``A child blowing bubbles that float and pop gently''}. These examples highlight the model's ability to generate high-quality videos with semantic and temporal coherence.

%% file: sec/0_abstract.tex
\begin{abstract}
Zero-shot, training-free, image-based text-to-video generation is an emerging area that aims to generate videos using existing image-based diffusion models. Current methods in this space require specific architectural changes to image-generation models, which limit their adaptability and scalability. In contrast to such methods, we provide a model-agnostic approach. We use intersections in diffusion trajectories, working only with the latent values. We could not obtain localized frame-wise coherence and diversity using only the intersection of trajectories. Thus, we instead use a grid-based approach. An in-context trained LLM is used to generate coherent frame-wise prompts; another is used to identify differences between frames. Based on these, we obtain a CLIP-based attention mask that controls the timing of switching the prompts for each grid cell. Earlier switching results in higher variance, while later switching results in more coherence. Therefore, Our approach can ensure appropriate control between coherence and variance for the frames. Our approach results in state-of-the-art performance while being more flexible when working with diverse image-generation models. The empirical analysis using quantitative metrics and user studies confirms our model’s superior temporal consistency, visual fidelity and user satisfaction, thus providing a novel way to obtain training-free, image-based text-to-video generation. Further examples and code at \href{https://djagpal02.github.io/EIDT-V/}{https://djagpal02.github.io/EIDT-V/}

\end{abstract}

%% file: sec/_1_Introduction.tex
\section{Introduction}

Recent advances in image generation have established diffusion models as state-of-the-art (SOTA) tools for producing visually compelling and coherent images. Open source techniques such as Stable Diffusion \cite{rombach_high-resolution_2022, podell_sdxl_2023, esser_scaling_2024} and Flux \cite{noauthor_flux_2024} show efficient high-quality outputs made possible by the diffusion models~\cite{song_score-based_2021, karras_elucidating_2022}. Despite this, extending these capabilities to video remains challenging due to the unique temporal coherence and dynamic motion requirements. 

SOTA, such as OpenAI's Sora \cite{noauthor_sora_2024} and Meta's MovieGen \cite{polyak_movie_2024}, have made significant progress in high-quality text-to-video generation. However, they rely on extensive training and complex architectures, which limits accessibility. These models typically require high-end GPUs or sit behind paywalls, making them out of reach for most users. Although lower cost models \cite{ho_video_2022, zhou_magicvideo_2023, ho_imagen_2022, gupta_photorealistic_2023, he_latent_2023, blattmann_align_2023, blattmann_stable_2023, gu_seer_2024, yang_video_2023, liu_video-p2p_2023, an_latent-shift_2023, liu_ed-t2v_2023} exist, the difference in quality is significant, and the inference and training costs are still beyond the reach of most. Hence, research is also being pursued to explore training-free approaches \cite{khachatryan_text2video-zero_2023, hong_direct2v_2024, huang_free-bloom_2023}. These approaches require architectural adjustments tied to specific diffusion models, limiting flexibility and scalability while producing even lower-quality outputs.

%\subsection{Proposed Approach}
We propose a model-agnostic, zero-shot text-to-video generation approach using image-based diffusion models. Our method operates entirely in the latent space, achieving compatibility with various image-based diffusion models without modifications or additional training. This model-agnostic design enables high-quality video generation across different architectures, establishing a flexible and robust foundation for video generation.

\textbf{Contributions:} The main contributions of our work are:
\begin{itemize}
\item \textbf{Grid-Based Prompt Switching for Conditional Image Generation:} We develop a novel grid-based prompt switching technique specifically for conditional image generation. This method divides each image into localized grid regions, providing fine-grained control over variance. 
\item \textbf{Enhanced Text control via LLM Modules :} Our approach incorporates two in-context trained modules within a LLM: one module generates frame-wise prompts from text inputs, while the other detects differences between consecutive frames to guide temporal consistency. 
\item \textbf{CLIP-based attention masking} - The text-guided difference between consecutive frames is used through a CLIP-based attention masking module that generates the prompt switch times, thereby controlling the variance and coherence spatially and between frames. 
\end{itemize}

All these contributions result in a complete pipeline thoroughly validated compared with corresponding baselines and ablations. We provide more comprehensive metrics to validate the generated results' temporal consistency and visual fidelity, along with suitable user studies.

%% file: sec/_2_RelatedWorks.tex
\section{Related Works}

\subsection{Image Diffusion}
Diffusion models \cite{sohl-dickstein_deep_2015, ho_denoising_2020, dhariwal_diffusion_2021, song_score-based_2021, song_denoising_2022} have become essential for generating high-quality images, offering improved stability and scalability over previous SOTA GANs \cite{goodfellow_generative_2014}. With continuous advancements \cite{song_consistency_2023, liu_flow_2022}, diffusion models are evolving rapidly. The open-source Stable Diffusion (SD) models illustrate this progression: moving from SD1 \cite{rombach_high-resolution_2022} to SDXL \cite{podell_sdxl_2023}, the architecture became significantly more complex. In SD3 \cite{esser_scaling_2024}, the original UNet architecture \cite{ronneberger_u-net_2015} was replaced by a multi-modal version of the diffusion image transformer \cite{peebles_scalable_2023} and enhanced with rectified flows \cite{liu_flow_2022}. These rapid architectural shifts often render tools for earlier versions obsolete, underscoring the need to develop adaptable tools.

\subsection{State-of-the-Art Video Generation Models}
SOTA video generation models, such as Runway’s Gen-3 Alpha \cite{noauthor_runway_nodate}, Meta’s MovieGen \cite{polyak_movie_2024}, and others \cite{noauthor_kling_nodate}, \cite{noauthor_sora_2024}, have demonstrated high-quality video synthesis with solid temporal coherence. However, they are costly to train and for inference, highlighting the need for computationally efficient alternatives.

\subsection{Low Cost Video Generation}
Initial efforts, such as VDM \cite{ho_video_2022} and MagicVideo \cite{zhou_magicvideo_2023}, adapted image diffusion processes to video, with MagicVideo leveraging latent spaces to lower computational requirements. Multi-stage models, like Imagen Video \cite{ho_imagen_2022}, utilize cascaded sub-models to enhance video resolution and frame rate, but the complexity of these stages keeps computational demands high. Similarly, transformer-based approaches, such as WALT \cite{gupta_photorealistic_2023}, employ memory-efficient windowed attention, though maintaining high-resolution coherence across frames still requires substantial resources.

Models like LVDM \cite{he_latent_2023} and Video LDM \cite{blattmann_align_2023} use hierarchical and latent diffusion techniques to extend video length, aiming for efficiency but still facing scalability challenges. Even efficiency-focused methods like Latent-Shift \cite{an_latent-shift_2023} and Ed-t2v \cite{liu_ed-t2v_2023} that enhance motion fidelity with techniques such as temporal shift modules and identity attention remain relatively expensive, making truly low-cost, accessible video generation a continuing challenge.

\subsection{Zero-Shot Video Generation}
Zero-shot video generation uses pre-trained image models for video synthesis tasks without additional training, enhancing accessibility. Text2Video-Zero \cite{khachatryan_text2video-zero_2023} was an early example, utilizing approximated optical flow and replacing self-attention with cross-attention to maintain frame continuity, although its limited motion approximation reduces scalability. Subsequent models such as Free-Bloom \cite{huang_free-bloom_2023} and DirecT2V \cite{hong_direct2v_2024} improved semantic coherence by using LLMs to generate frame-wise prompts. However, the LLMs in these models were manually configured and lacked standardized frameworks, which affects reproducibility. Moreover, these approaches are heavily tied to specific diffusion architectures, making them susceptible to becoming obsolete as diffusion models evolve.

%% file: sec/_3_Background.tex
\section{Background}

\subsection{Foundational Concepts: Diffusion Models}

Diffusion models transform an initial noise sample \( X_T \) into a final data sample \( X_0 \) through iterative denoising over a time horizon \( T \). Two main approaches are denoising diffusion probabilistic models (DDPMs) \cite{sohl-dickstein_deep_2015,ho_denoising_2020}, which progressively remove noise in discrete steps, and score-based models \cite{song_generative_2019}, which leverage gradients of data densities to guide the denoising process. Karras et al. \cite{karras_elucidating_2022} unified these methods into a modular framework, showing that both aim to map noise to data through a continuous denoising trajectory.

By treating the diffusion steps as continuous over \( t \in [0, T] \), Song et al. \cite{song_score-based_2021} reformulated diffusion as Stochastic Differential Equations (SDEs), where the reverse SDE describes the process of recovering data from noise. They further introduced Probability Flow ODEs (PF-ODEs) \cite{song_score-based_2021}, which offer a deterministic path from noise to data while preserving the identical marginal distributions as the SDE. This continuous, deterministic trajectory simplifies the sampling process, making diffusion models efficient for high-quality image generation.

\subsection{Classifier-Free Guidance}

Classifier-free guidance \cite{ho_classifier-free_2022} enables conditioning diffusion models on text without relying on an external classifier. During training, the model alternates between a specific condition \( y \) (e.g., text) and a null condition \( \emptyset \), learning both the conditional \( s_{\theta}(x_t, t \mid y) \) and unconditional \( s_{\theta}(x_t, t \mid \emptyset) \) score functions. The final conditional score function at sampling is given by:
\begin{align}
    \nabla_{x_t} \log p_{X_t|Y}(x_t|y) &\approx s_{\theta}(x_t, t \mid y) \notag \\
    &\quad + \gamma \left( s_{\theta}(x_t, t \mid y) - s_{\theta}(x_t, t \mid \emptyset) \right),
\end{align}
where \(\gamma\) controls the influence of the text condition. Adjusting the guidance scale, \(\gamma\), allows the diffusion process to produce images that align closely with the text prompt.

\subsection{Uniqueness of Diffusion Trajectories}

In the deterministic framework of ODE-based diffusion, each trajectory—initiated from a state \( x_T \) under a given condition \( y \)—uniquely determines the final output \( x_0 \), provided the ODE satisfies Lipschitz continuity \cite{song_score-based_2021, karras_elucidating_2022}. This property ensures that diffusion models using ODE formulations yield consistent trajectories for a given condition.

%% file: sec/_4_Methodology.tex
\section{Methodology}
We present the EIDT-V pipeline, illustrated in \cref{fig:pipeline}. This pipeline leverages diffusion intersections to enable frame-based video generation. The following subsections detail each component, beginning with the foundational intuition behind our approach.

\input{Figures/pipeline}

\subsection{Prompt Switching via Diffusion Intersection}

Consider two vehicles, \( A \) and \( B \), each travelling backwards in time from an initial time \( T \)  toward a destination at time \( 0 \). The position of vehicle \( i \) at time \( t \) is denoted by \( \mathbf{x}^{(i)}_t \), where \( i \in \{A, B\} \), and each vehicle follows a deterministic trajectory:

\begin{equation}
\mathbf{x}^{(i)}_t = \mathbf{f}(\mathbf{x}^{(i)}_T, t, y_i),
\end{equation}

where \( \mathbf{x}^{(i)}_T \) is the starting position, and \( y_i \) specifies the guidance or ``route" for vehicle \( i \). 

If the trajectories of \( A \) and \( B \) intersect at time \( t = t_s \), a dependency forms, constraining their maximum separation at the destination, \( t = 0 \). Assuming each vehicle moves with speed \( v \), this maximum separation at \( t = 0 \) is given by the distance function \( D(t_s) = 2 \cdot v \cdot t_s \), such that:

\begin{equation}
\| \mathbf{x}^{(A)}_0 - \mathbf{x}^{(B)}_0 \| \leq D(t_s).
\end{equation}

In diffusion models, this analogy applies to image synthesis, where each trajectory represents the ODE evolution of an image from an initial noisy state \( x_T \) to a coherent structure \( x_0 \), guided by a prompt \( y \). By switching the Prompt from \( y \) to \( y' \) at time \( t = t_s \), we create a similar intersection point that limits divergence between the resulting images, allowing us to control how much each prompt influences the final image. 

\subsection{Grid Prompt Switching}

Prompt switching provides global variance control; however, in some cases, we may want more targeted variance. To address this, we introduce \textbf{grid prompt switching}. In this method, we divide the image into an \( n \times n \) grid, where each cell \((i, j)\) is assigned its own prompt switch time \( t_s^{(i, j)} \). This split allows specific image regions to adopt prompt changes independently, enabling fine-grained control over which parts of the image transition to a new prompt.

Prompt switching for each grid cell individually can lead to inconsistencies, as the diffusion process is inherently global and interconnected throughout the image. We propose a hybrid approach combining the original diffusion trajectory with the updated one to maintain spatial coherence while allowing for localized prompt transitions.

As illustrated in \cref{fig:grid_prompt_switching}, we implement this by first defining a \textit{Switch Time Matrix}  (STM), which determines the specific time \( t_s^{(i, j)} \) at which each cell switches prompts. This matrix is compared with the current timestep \( t \) to create a binary mask \( M_t^{(i, j)} \) for each cell, signalling whether at time  \( t \) it should follow the original trajectory or denoise a new one.

\begin{equation}
\label{eq:mask}
M_t^{(i, j)} = 
\begin{cases}
0, & \text{if } t > t_s^{(i, j)} \\[6pt]
1, & \text{if } t \leq t_s^{(i, j)}
\end{cases}
\end{equation}

At each diffusion step \( t \), the mask \( M_t \) dynamically determines which parts of the image should use latents from the original trajectory \( X_t^{(A)} \) and which should be updated with latents from the new prompt trajectory \( X_t^{(B)} \). The latent
representation for each cell is, therefore, given by:

\begin{equation}
\label{eq:combined_latent}
X_t^{(i,j)} = 
\begin{cases}
X_t^{(A, i, j)}, & \text{if } M_t^{(i, j)} = 0 \\[6pt]
X_t^{(B, i, j)}, & \text{if } M_t^{(i, j)} = 1
\end{cases}
\end{equation}

This results in a composite latent \( X_t \) for the image, calculated as follows:

\begin{equation}
\label{eq:composite_refined}
X_t = M_t \odot X_t^{(B)} + (1 - M_t) \odot X_t^{(A)}
\end{equation}

Where \( \odot \) denotes element-wise multiplication applied across each cell in the latent representation.

This technique preserves overall image coherence by seamlessly transitioning cells between prompts, effectively combining the inpainting and generation processes.

\subsection{Text-Guided Grid Switching with Attention}

\input{Figures/grid_prompt_switching}

Selecting an effective STM is essential to ensure that prompt transitions align with areas of significant variation. As shown in \cref{fig:grid_prompt_switching} (right panel), we automate this by comparing the initial prompt \( y \) with the target prompt \( y' \) to obtain a textual difference \( \Delta\). This difference highlights regions of high attention, where prompt \( y' \) introduces changes or motion.

Given the list of differences, expected differences in \cref{fig:grid_prompt_switching}, we apply a CLIP-Segmentation model~\cite{luddecke_image_2022} to create attention maps over the image \( x_0^{A} \). The attention maps are normalized, exponentiated, and resized to match the latent size, producing the targeted STM. Cells with high attention values receive earlier switch times, allowing more variance, while low-attention areas retain stability.

As illustrated in \cref{fig:pipeline}, this approach connects frames in the video module, enabling controlled transitions in motion-heavy areas for smooth frame continuity.

\subsection{Large Language Models}

We now have a mechanism to connect frames in our video, but we require frame-wise conditional text prompts and lists of differences between them. To generate these, we use in-context learning~\cite{brown_language_2020, min_rethinking_2022} with LLaMA-3 8B~\cite{dubey_llama_2024}, creating two modules as shown in \cref{fig:pipeline}.

\textbf{LLM Framewise Prompts} generates prompts \( y_t \) for each frame \( t \) based on the user’s initial description. Each Prompt consists of a fixed scene descriptor and a dynamic component:

\begin{equation}
y_t = \text{Fixed Scene Descriptor} + \text{Dynamic Component}_t
\label{eq:framewise_prompt}
\end{equation}

To fit within CLIP's~\cite{radford_learning_2021} 77-token limit, we allocate 60 tokens for the fixed part and 15 for the dynamic component (2-token buffer), preserving prompt coherence across frames while allowing gradual transitions.

\textbf{LLM Difference Detector} compares two text prompts to identify differences, returning a list of distinct elements between them. We ask the model to return anything that may be in motion or varied between the two scenes described by the prompts.

Examples of outputs from both modules are in the text module in \cref{fig:pipeline} and in \cref{fig:compact_attention_examples}.

%% file: Figures/pipeline.tex
\begin{figure*}[ht]
    \centering
    \includegraphics[width=0.8\textwidth]{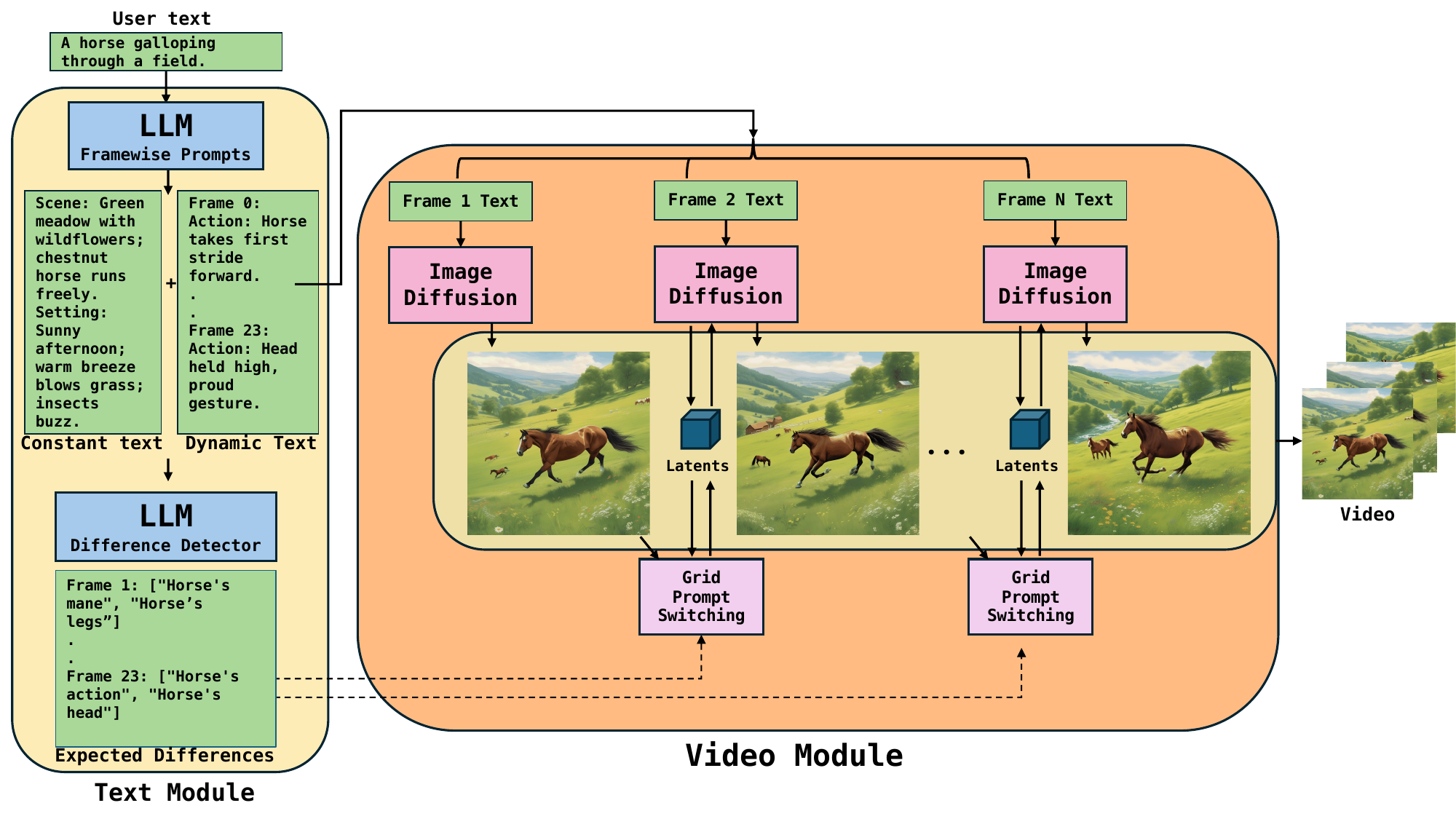}
    \caption{\textbf{EIDT-V Pipeline for Frame-Based Video Generation.} The pipeline consists of two primary modules: text and video. The text module converts the user’s input into framewise prompts and expected variations, which guide the video module in generating frames iteratively. The video module achieves controlled variance and coherence across frames by leveraging trajectory intersections. Integrating two LLM modules and the grid prompt switching enables a generic image diffusion model to synthesize coherent video sequences effectively.}
    \label{fig:pipeline}
\end{figure*}

%% file: Figures/grid_prompt_switching.tex
\begin{figure}[ht]
    \centering
    \includegraphics[width=0.9\columnwidth]{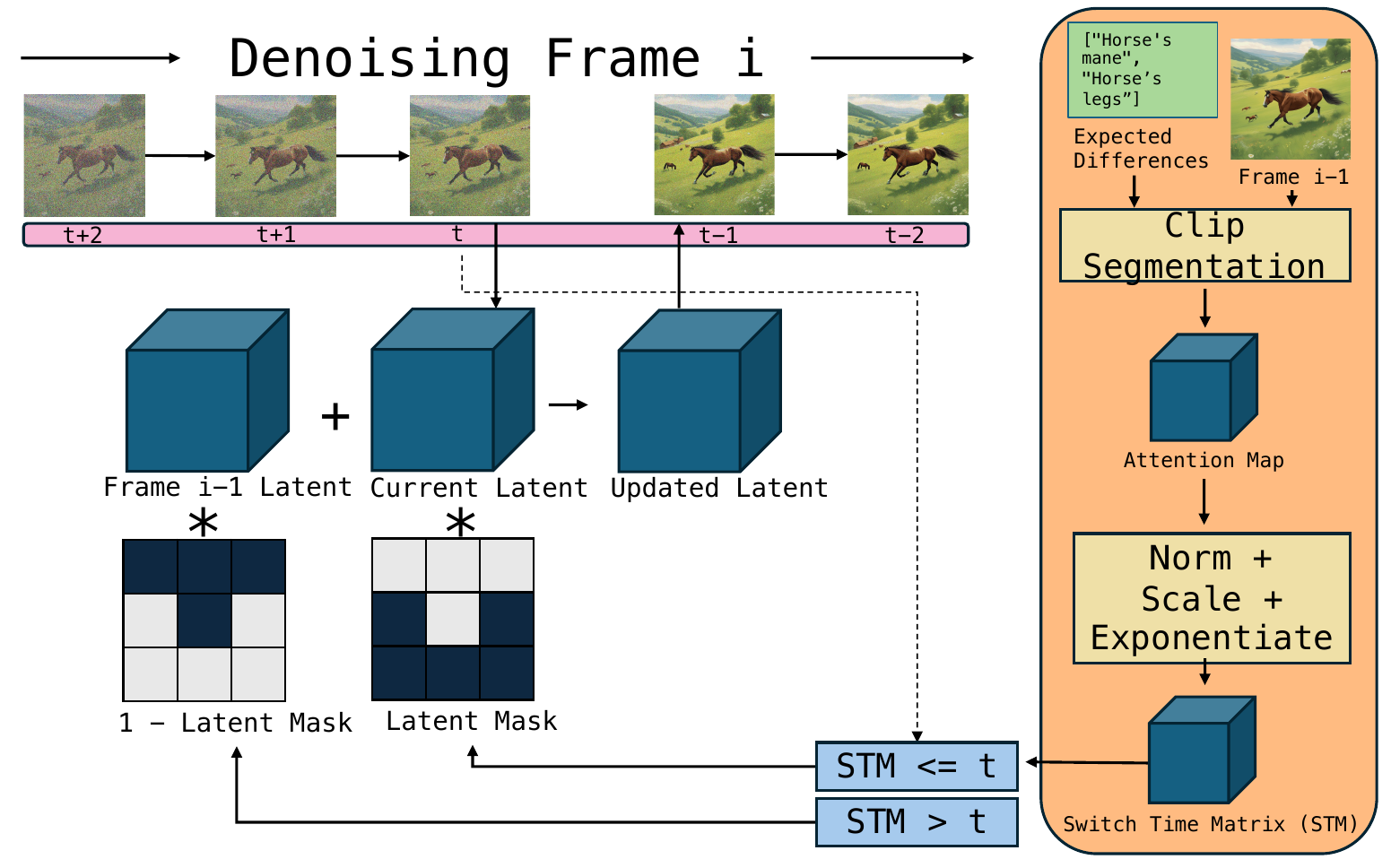}
    \caption{\textbf{Grid Prompt Switching with Text-Guided Attention.} On the right, an auxiliary block processes the previous frame and difference text using a CLIP segmentation model to generate the STM. This is converted into a binary mask (\cref{eq:mask}) that selects, at timestep \(t\), whether each grid cell follows the original or new prompt trajectory. In the main denoising process, the mask blends the latent representations \(X_t^{(A)}\) and \(X_t^{(B)}\) as per \cref{eq:composite_refined} to form \(X_t\). While a coarse \(3 \times 3\) grid is shown for clarity, in practice, higher-resolution masks (e.g., \(128 \times 128\)) are used.}
    \label{fig:grid_prompt_switching}
\end{figure}

%% file: sec/_5_Results.tex
\section{Experiments}
\label{sec:experiments}

\subsection{Implementation Details}
\label{sec:implementation}

To evaluate our method's robustness, we created a set of 50 diverse video generation prompts (see \cref{sup:test_prompts}) selected to span various visual and semantic contexts.

We implement our model with the Stable Diffusion framework, primarily SD1.5~\cite{rombach_high-resolution_2022} for comparability with prior work. We also tested compatibility and performance on more advanced architectures, specifically SDXL~\cite{podell_sdxl_2023} and SD3 Medium~\cite{esser_scaling_2024}. All experiments ran on a single NVIDIA RTX A5000 GPU (24 GB memory), comparable to an RTX 3090, demonstrating the feasibility of our approach on consumer-grade hardware. Hyperparameter tuning details can be found in \cref{sup:hyperparameter_selection}.

We also explored a modular alternative to cross-attention manipulation of previous works using the IP-Adapter~\cite{ye_ip-adapter_2023}, more details in \cref{sup:ip_adapter}. 

\subsection{Evaluation Metrics}
\label{sec:evaluation_metrics}

Previous works have often relied on the CLIP score as a primary evaluation tool; however, this method only checks image-text alignment. To address this gap, we incorporated additional metrics specifically designed to assess frame-to-frame consistency. First, we employed Multi-Scale Structural Similarity (MS-SSIM) \cite{wang_multiscale_2003}, quantifying the structural similarity between consecutive frames across multiple scales. Higher MS-SSIM values indicate better preservation of structural information, which is essential for maintaining consistent content and layout between frames. Additionally, we used Learned Perceptual Image Patch Similarity (LPIPS) \cite{zhang_unreasonable_2018}, which assesses perceptual similarity by analyzing deep features from neural networks. Lower LPIPS values correlate with better perceptual continuity across frames, contributing to smoother transitions and reducing flickering artefacts. Finally, we introduced an Optical Flow-based Temporal Consistency Loss using the Farneback method \cite{farneback_two-frame_2003} to evaluate temporal motion consistency. Lower temporal consistency loss values suggest smoother, more coherent motion, an essential factor in generating realistic video sequences. These metrics provide a more comprehensive assessment of our method's capacity to generate videos with high temporal coherence, structural integrity, and perceptual consistency across frames. For completeness, we report CLIP scores in \cref{sup:clip_results}, although, as expected, these show minimal variance across different models.

\subsection{Results}
\label{sec:results}

This section presents quantitative and qualitative evaluations of our method, comparing its performance to baseline approaches under two conditions: (1) using Stable Diffusion 1.5 for fair comparisons with existing works and (2) evaluating its adaptability and scalability across different architectures. We emphasize that our comparisons are limited to training-free methods with publicly available code, as many more recent approaches require additional training or lack accessible implementations.

\input{Figures/main_qualitative}

\subsubsection{Quantitative Comparison}
\label{sec:quantitative}

Our method achieved competitive performance across all three metrics compared to previous approaches using SD1.5. Specifically, our MS-SSIM score for EIDT-V SD1.5 with IP-Adapter was $0.655 \pm 0.13$, closely matching the highest score of $0.672 \pm 0.095$ achieved by Free-Bloom \cite{huang_free-bloom_2023}. Furthermore, our LPIPS score of $0.316 \pm 0.089$ was lower than those obtained by other SD1.5-based methods, indicating an improvement in perceptual similarity across frames. Regarding temporal consistency, our base SD1.5 model achieved a score of $0.152 \pm 0.062$, suggesting that our model generated more stable motion as measured by optical flow analysis.

Our method demonstrated strong adaptability to newer frameworks when evaluating across architectures, with substantial performance improvements. For example, the SDXL \cite{podell_sdxl_2023} implementation produced an MS-SSIM score of $0.701 \pm 0.089$, with LPIPS and Temporal Consistency scores of $0.28 \pm 0.086$ and $0.138 \pm 0.054$, respectively. The SD3 Medium \cite{esser_scaling_2024} model further enhanced performance, achieving MS-SSIM, LPIPS, and Temporal Consistency scores of $0.81 \pm 0.109$, $0.184 \pm 0.08$, and $0.087 \pm 0.042$, respectively. These results underscore the scalability of our approach, with newer architectures contributing to improved detail, reduced flickering, and greater coherence.

\input{Tables/main_quantitative}

\subsubsection{Qualitative Comparison}
\label{sec:qualitative_comparison}

Qualitative comparisons with baseline methods \cref{fig:main_qual} (additional in \cref{sup:additional_qualitative}, \cref{sup:additional_best}), further illustrate our method's improvements in temporal coherence, subtle motion accuracy and flexibility across various architectures. 

When evaluated on the same SD1.5 architecture as prior methods such as DirecT2V \cite{hong_direct2v_2024}, Free-Bloom \cite{huang_free-bloom_2023}, and T2V-Zero \cite{khachatryan_text2video-zero_2023}, our approach demonstrated notable qualitative enhancements. Our method achieved smoother frame transitions and maintained consistency across sequential frames, resulting in visually coherent videos. Additionally, our approach effectively captured subtle, nuanced movements between frames, attributable to targeted attention adjustments applied across sequences.

We validated further on SDXL \cite{podell_sdxl_2023} and SD3 Medium \cite{esser_scaling_2024} architectures, where it produced frames with improved clarity and detail, enhancing the realism of video sequences. Notably, with high fidelity, our approach captured text-prompt-specific changes, such as colour variations or subtle movements, such as hand waves. These highlight our model's flexibility and capacity to adhere promptly across diverse video generation scenarios.

\subsection{User Study}
\label{sec:user_study}

To evaluate our approach from a human perspective, we conducted a user study comparing EIDT-V SD1.5 with three baseline models: FreeBloom, T2VZero, and DirecT2V. Eight participants assessed 50 video samples each across multiple criteria, including temporal coherence, visual fidelity, and semantic alignment. Detailed study design information is available in \cref{sup:user_study_setup}.

\input{Tables/user_study}

The results of the user study, shown in \cref{tab:userstudy}, indicate that our model, EIDT-V SD1.5, achieved the best scores across all but semantic coherence, where it was a close second. Our model received the best mean score of 1.9 for temporal coherence, indicating smoother transitions and improved frame-to-frame consistency. Regarding fidelity, our model achieved a mean score of 2.0, which is on par with FreeBloom, with users noting the superior visual quality and fewer artifacts than other models. For semantic coherence, EIDT-V scored 2.1, just shy of FreeBlooms 2.0, demonstrating strong alignment with the intended prompts. Our model received a mean score of 1.9 for user satisfaction, outperforming all baseline methods. Feedback shows that our model produces smoother video, focusing more on coherence, whereas FreeBloom focuses more on text alignment. 

\subsection{Ablation Study}
\label{sec:ablation}

To evaluate the impact of critical components in our approach, we conducted an ablation study analyzing how different configurations affect video generation quality across our metrics. Specifically, we tested configurations with and without Grid Prompt Switching (GrPS) and Our Framewise Prompts (OFP) to isolate their effects. The results of the ablation study are in \cref{tab:ablation}.

\input{Tables/ablation}

The results reveal that incorporating GrPS significantly improves structural and perceptual coherence in generated videos. For instance, the MS-SSIM score increases substantially when GrPS is added, rising from $0.132 \pm 0.052$ for the baseline ChatGPT (CG) prompts configuration to $0.588 \pm 0.129$ for CG + GrPS. Similarly, LPIPS, which measures perceptual similarity, decreases from $0.723 \pm 0.043$ in the CG configuration to $0.384 \pm 0.094$ in CG + GrPS, indicating reduced perceptual artifacts across frames. Temporal Consistency Loss also improves markedly, dropping from $0.389 \pm 0.046$ in CG to $0.157 \pm 0.062$, suggesting smoother and more coherent motion.

When OFP combines with GrPS (OFP + GrPS), the results are further enhanced, with MS-SSIM reaching $0.63 \pm 0.137$, LPIPS improving to $0.33 \pm 0.1$, and Temporal Consistency Loss reduced to $0.152 \pm 0.062$. These values represent the best performance across all configurations, confirming that the combination of framewise prompts and grid prompt switching provides the most consistent and visually coherent results.

\input{Figures/ablation_qual}

The qualitative analysis shown in \cref{fig:ablation} supports these findings, illustrating that without GrPS, frame-to-frame consistency is minimal. With GrPS, frame transitions become significantly smoother, and when paired with OFP, consistency is enhanced further, resulting in highly stable and coherent video sequences. Overall, the ablation study demonstrates the critical role of GrPS and OFP in achieving high-quality, temporally consistent video generation.

%% file: Figures/main_qualitative.tex
\begin{figure*}[t] % Use figure* to span both columns in a two-column layout
    \centering
    % Row for T2VZero images
    \begin{minipage}[t]{\textwidth}
        \centering
        \textbf{T2VZero}
    \end{minipage}
    \begin{minipage}[t]{0.32\textwidth}
        \centering
        \includegraphics[width=\linewidth]{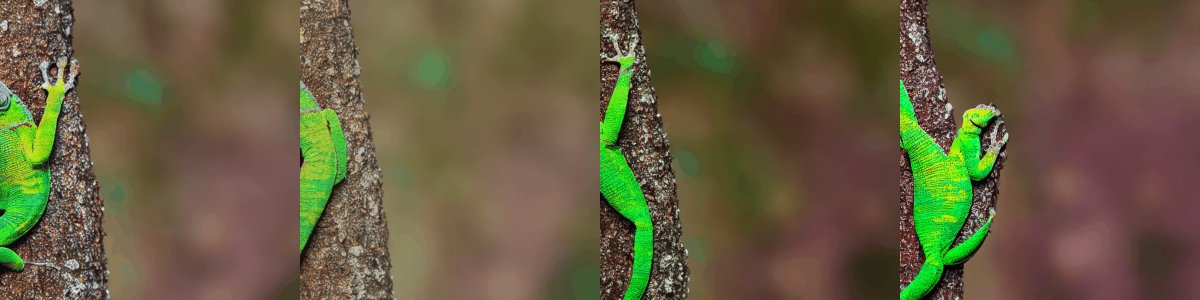}
    \end{minipage}
    \begin{minipage}[t]{0.32\textwidth}
        \centering
        \includegraphics[width=\linewidth]{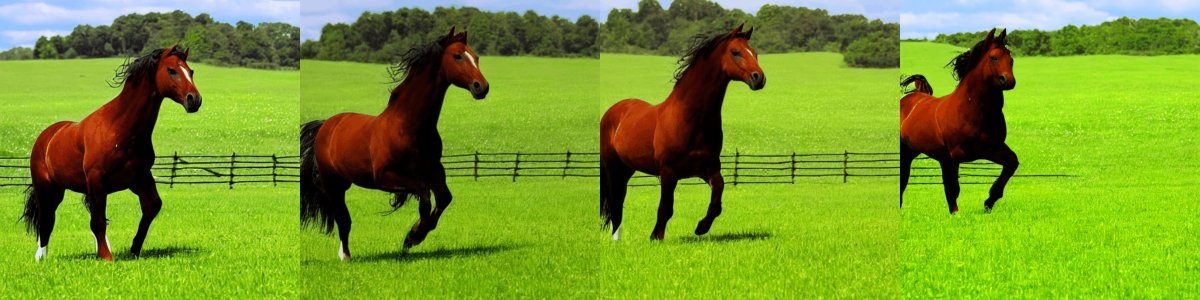}
    \end{minipage}
    \begin{minipage}[t]{0.32\textwidth}
        \centering
        \includegraphics[width=\linewidth]{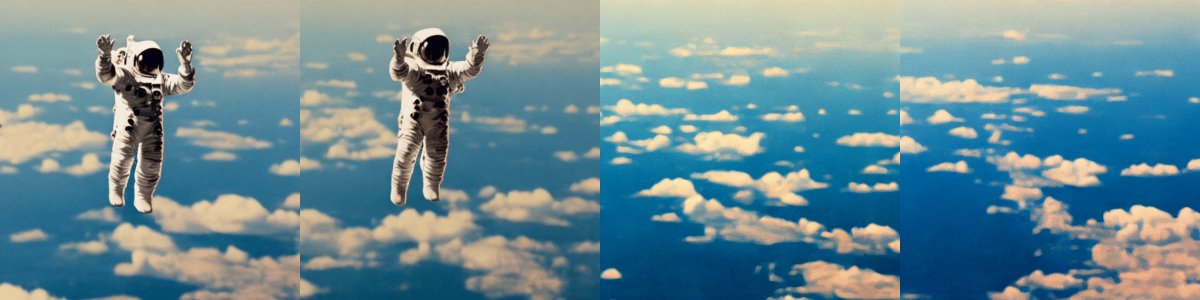}
    \end{minipage}

      % Space between image rows

    % Row for DirecT2V images
    \begin{minipage}[t]{\textwidth}
        \centering
        \textbf{DirecT2V}
    \end{minipage}
    \begin{minipage}[t]{0.32\textwidth}
        \centering
        \includegraphics[width=\linewidth]{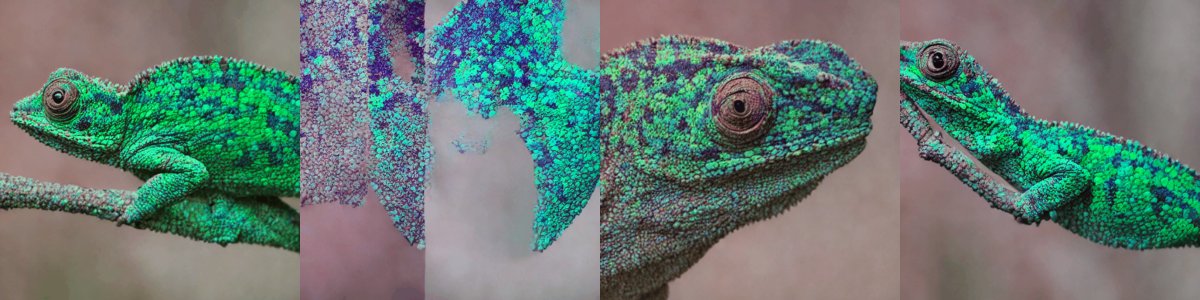}
    \end{minipage}
    \begin{minipage}[t]{0.32\textwidth}
        \centering
        \includegraphics[width=\linewidth]{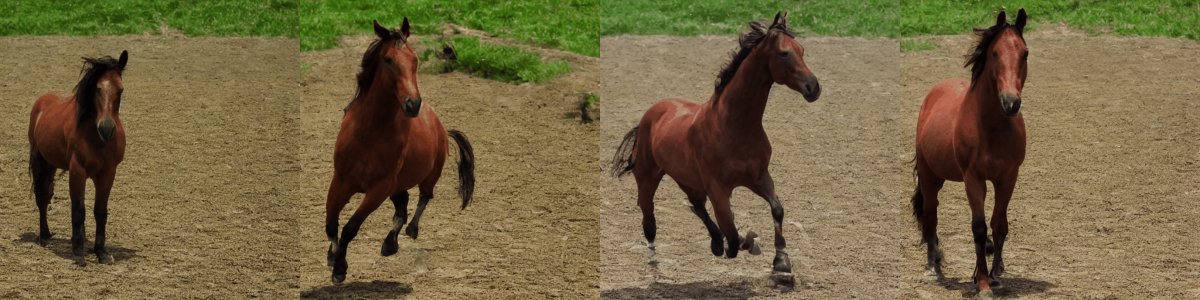}
    \end{minipage}
    \begin{minipage}[t]{0.32\textwidth}
        \centering
        \includegraphics[width=\linewidth]{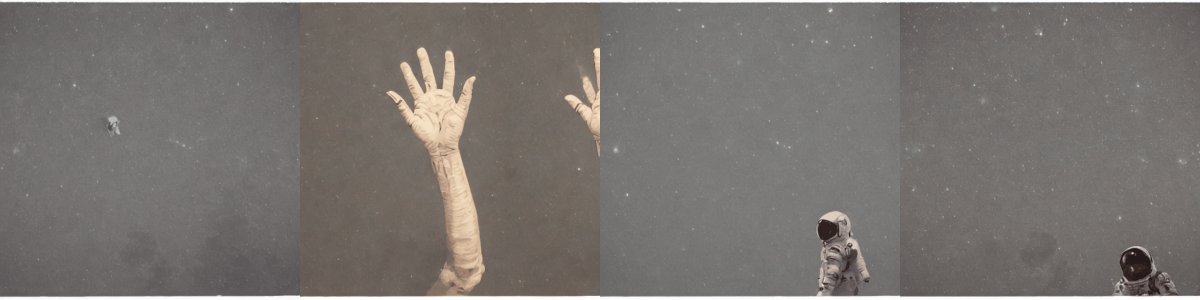}
    \end{minipage}

    % Row for FreeBloom images
    \begin{minipage}[t]{\textwidth}
        \centering
        \textbf{FreeBloom}
    \end{minipage}
    \begin{minipage}[t]{0.32\textwidth}
        \centering
        \includegraphics[width=\linewidth]{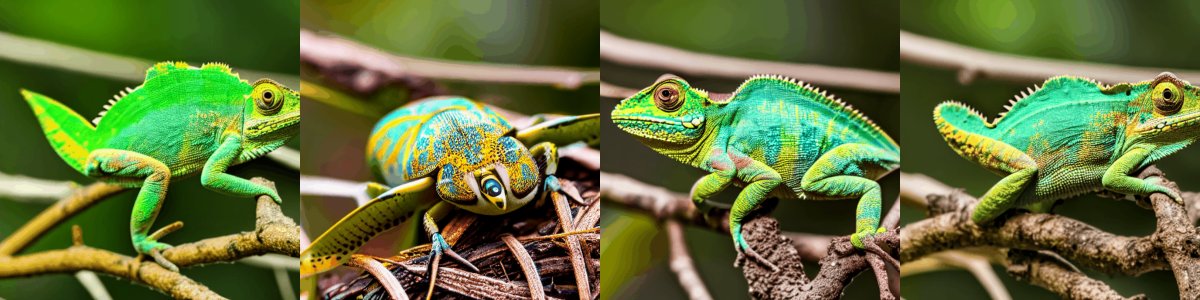}
    \end{minipage}
    \begin{minipage}[t]{0.32\textwidth}
        \centering
        \includegraphics[width=\linewidth]{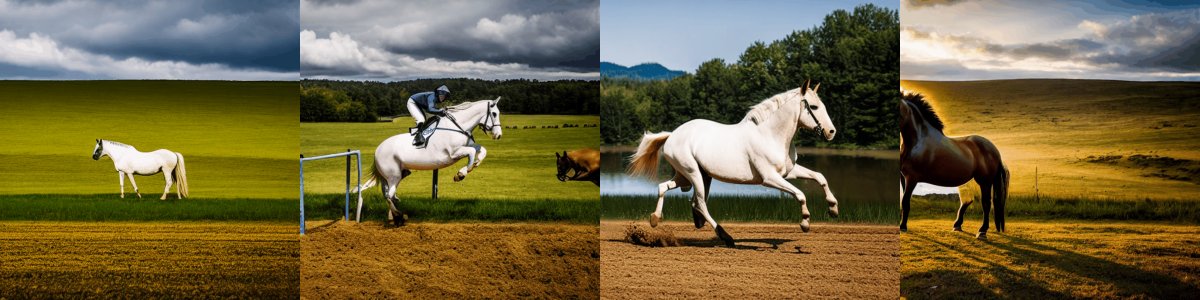}
    \end{minipage}
    \begin{minipage}[t]{0.32\textwidth}
        \centering
        \includegraphics[width=\linewidth]{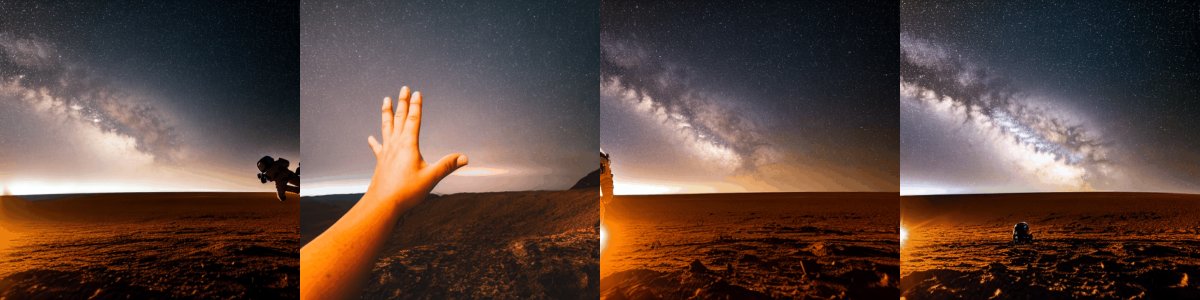}
    \end{minipage}

    % Row for EIDT-V SD images
    \begin{minipage}[t]{\textwidth}
        \centering
        \textbf{EIDT-V SD}
    \end{minipage}
    \begin{minipage}[t]{0.32\textwidth}
        \centering
        \includegraphics[width=\linewidth]{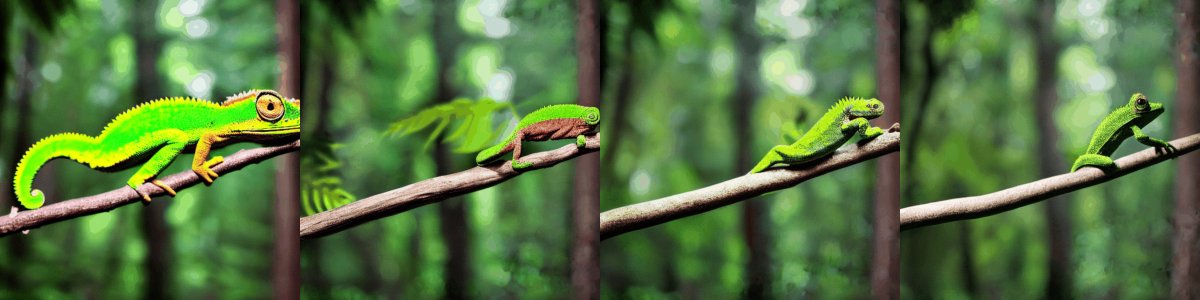}
    \end{minipage}
    \begin{minipage}[t]{0.32\textwidth}
        \centering
        \includegraphics[width=\linewidth]{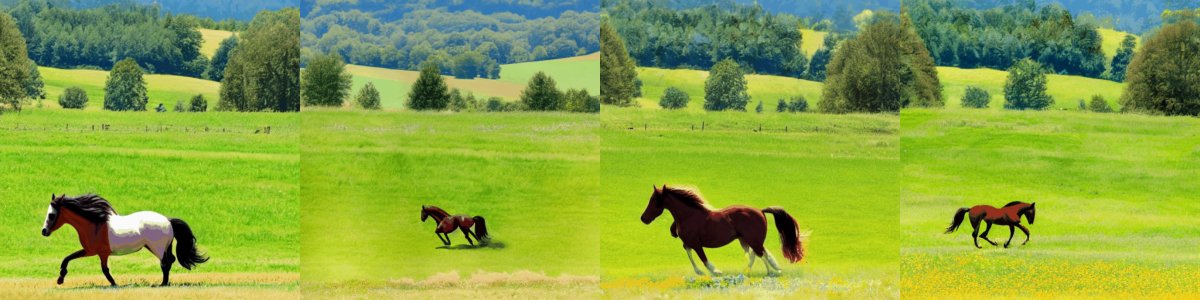}
    \end{minipage}
    \begin{minipage}[t]{0.32\textwidth}
        \centering
        \includegraphics[width=\linewidth]{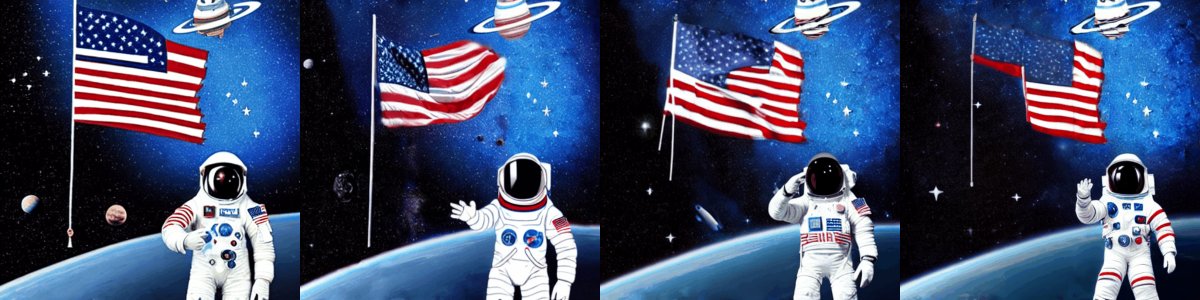}
    \end{minipage}

    % Row for EIDT-V SD_IP images
    \begin{minipage}[t]{\textwidth}
        \centering
        \textbf{EIDT-V SD\_IP}
    \end{minipage}
    \begin{minipage}[t]{0.32\textwidth}
        \centering
        \includegraphics[width=\linewidth]{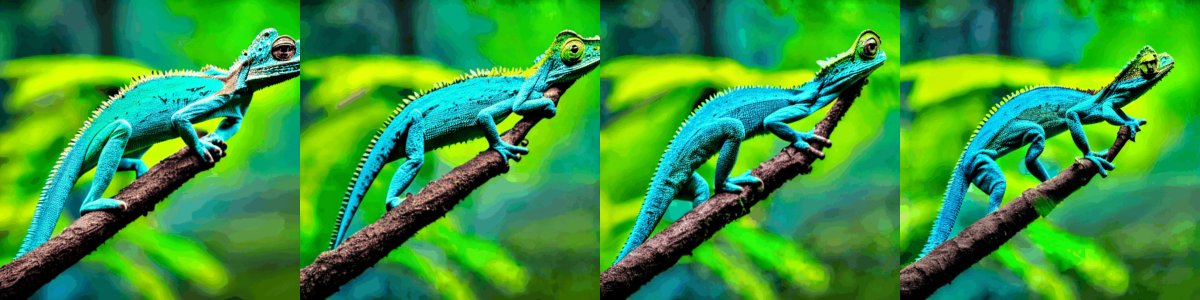}
    \end{minipage}
    \begin{minipage}[t]{0.32\textwidth}
        \centering
        \includegraphics[width=\linewidth]{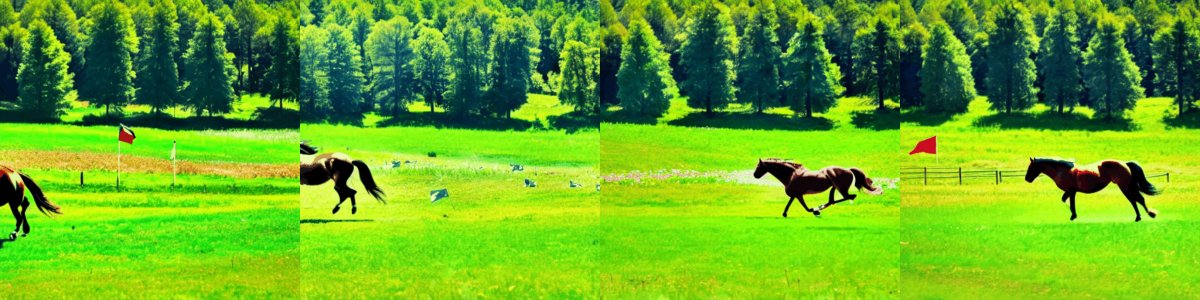}
    \end{minipage}
    \begin{minipage}[t]{0.32\textwidth}
        \centering
        \includegraphics[width=\linewidth]{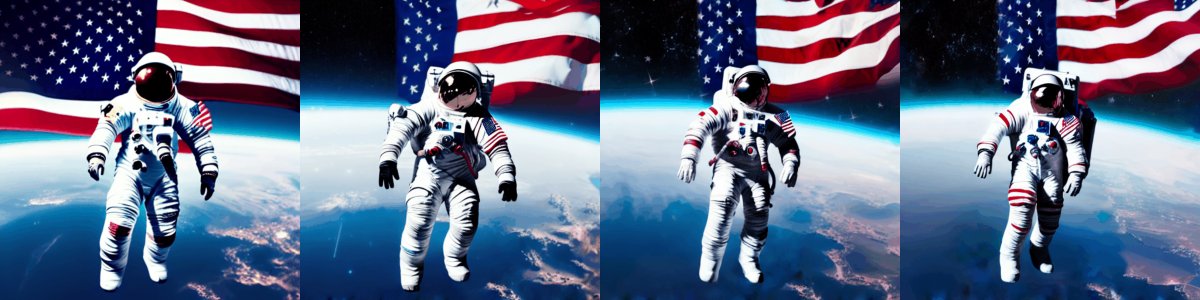}
    \end{minipage}

    % Row for EIDT-V SDXL images
    \begin{minipage}[t]{\textwidth}
        \centering
        \textbf{EIDT-V SDXL}
    \end{minipage}
    \begin{minipage}[t]{0.32\textwidth}
        \centering
        \includegraphics[width=\linewidth]{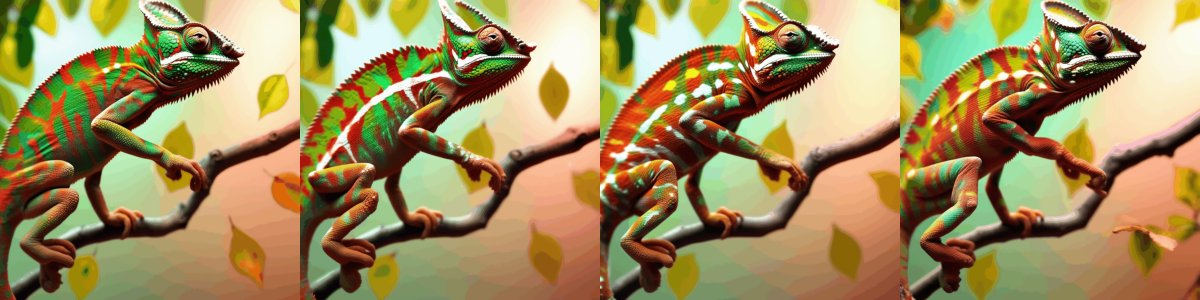}
    \end{minipage}
    \begin{minipage}[t]{0.32\textwidth}
        \centering
        \includegraphics[width=\linewidth]{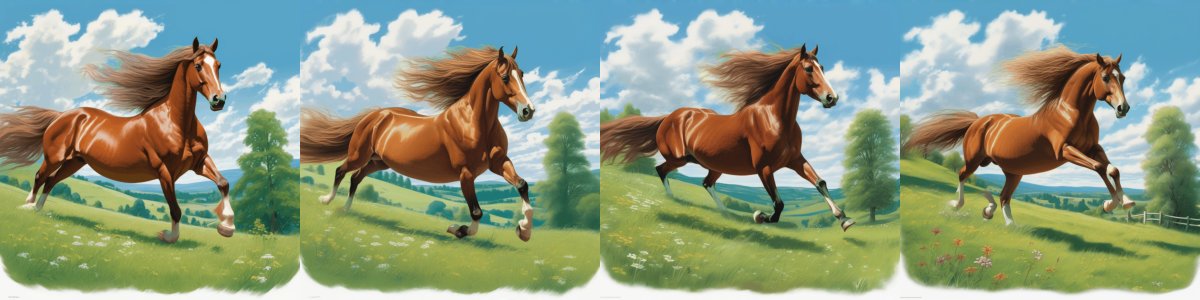}
    \end{minipage}
    \begin{minipage}[t]{0.32\textwidth}
        \centering
        \includegraphics[width=\linewidth]{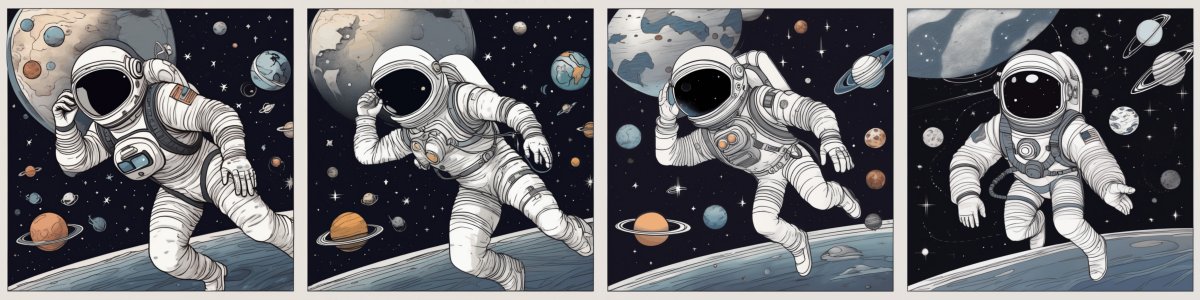}
    \end{minipage}

    % Row for EIDT-V SD3 images
    \begin{minipage}[t]{\textwidth}
        \centering
        \textbf{EIDT-V SD3}
    \end{minipage}
    \begin{minipage}[t]{0.32\textwidth}
        \centering
        \includegraphics[width=\linewidth]{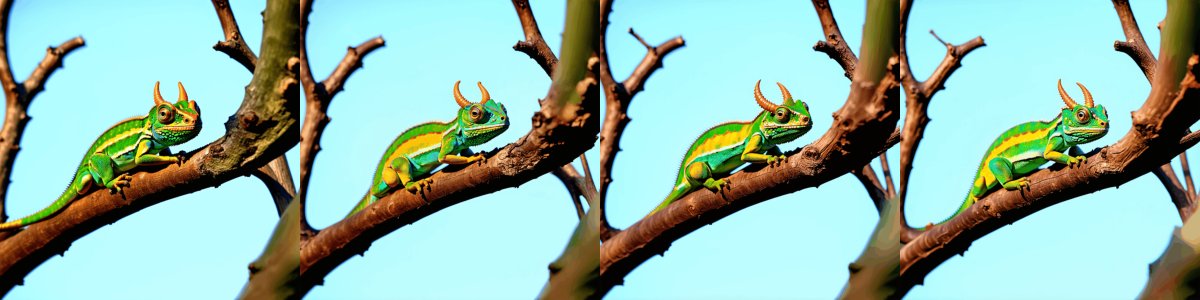}
    \end{minipage}
    \begin{minipage}[t]{0.32\textwidth}
        \centering
        \includegraphics[width=\linewidth]{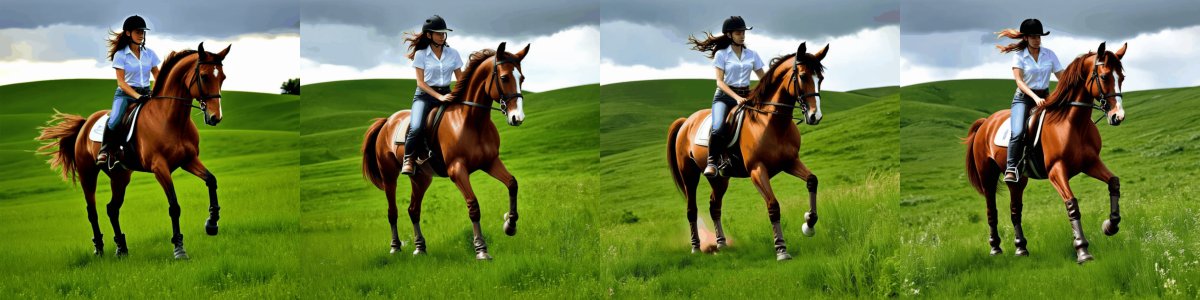}
    \end{minipage}
    \begin{minipage}[t]{0.32\textwidth}
        \centering
        \includegraphics[width=\linewidth]{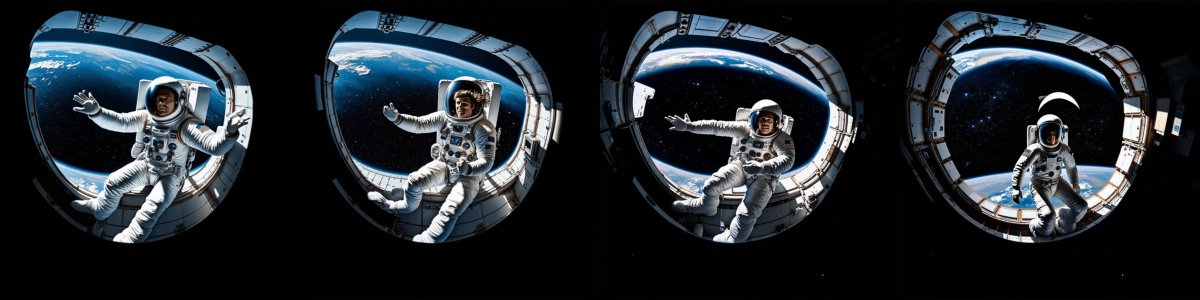}
    \end{minipage}

    \caption{\textbf{Qualitative comparison of different video generation models across three prompts:} (a) ``\textbf{A chameleon changing colors on a branch}", (b) ``\textbf{A horse galloping across a field}", and (c) ``\textbf{An astronaut floating in space waving}". T2VZero produces coherent frames but does not fully capture the specifics of each prompt; for instance, the chameleon does not change colors, and the astronaut does not appear to be waving. DirecT2V struggles to generate coherent frames. Interestingly, both DirecT2V and FreeBloom, which are LLM-based models, capture the essence of ``waving" and "space" but fail to fully integrate these concepts in each frame. They have strong semantic coherence but not temporal. Our model, however, demonstrates clear color changes in the chameleon, captures the horse's movement (notice the legs), and shows the astronaut's arm moving in a waving pattern while keeping the rest of the frame highly consistent.}
    \label{fig:main_qual}
\end{figure*}

%% file: Tables/main_quantitative.tex
\begin{table*}[t]
\centering
\caption{\textbf{Quantitative performance comparison of video generation models}, including T2V-Zero, DirecT2V, Free-Bloom, and our proposed EIDT-V method across multiple configurations. Metrics include MS-SSIM (higher indicates better structural similarity), LPIPS (lower indicates better perceptual quality), and Temporal Consistency Loss (lower indicates better temporal coherence). The table highlights the flexibility of EIDT-V across various pre-trained architectures, with the best results achieved using SD3 Medium.}
\label{tab:quantitative}
\resizebox{0.8\linewidth}{!}{%
\begin{tabular}{ll|c|ccc}
\toprule
\textbf{Method} & \textbf{Pre-Trained} & \textbf{Unmodified} & \textbf{MS-SSIM} & \textbf{LPIPS} & \textbf{Temporal} \\ 
 & \textbf{Model} & \textbf{Architecture} & ($\uparrow$) & ($\downarrow$) & \textbf{Consistency} ($\downarrow$) \\ 
\midrule
\textbf{T2V-Zero \cite{khachatryan_text2video-zero_2023}} & \textbf{SD1.5} & \xmark & $0.428 \pm 0.174$ & $0.404 \pm 0.083$ & $0.206 \pm 0.066$ \\ 
\textbf{DirecT2V \cite{hong_direct2v_2024}} & \textbf{SD1.5} & \xmark & $0.492 \pm 0.135$ & $0.445 \pm 0.089$ & $0.185 \pm 0.061$ \\ 
\textbf{Free-Bloom \cite{huang_free-bloom_2023}} & \textbf{SD1.5} & \xmark & $\mathbf{0.672 \pm 0.095}$ & $0.353 \pm 0.082$ & $0.159 \pm 0.039$ \\ 
\textbf{EIDT-V} & \textbf{SD1.5} & \cmark & $0.63 \pm 0.137$ & $0.33 \pm 0.1$ & $\mathbf{0.152 \pm 0.062}$ \\ 
\textbf{EIDT-V} & \textbf{SD1.5 w/ IP-Adapter \cite{liu_flow_2022}} & \xmark & $0.655 \pm 0.13$ & $\mathbf{0.316 \pm 0.089}$ & $0.158 \pm 0.074$ \\ 
\midrule
\textbf{EIDT-V} & \textbf{SDXL} & \cmark & $0.701 \pm 0.089$ & $0.28 \pm 0.086$ & $0.138 \pm 0.054$ \\ 
\textbf{EIDT-V} & \textbf{SD3 Medium} & \cmark & $\mathbf{0.81 \pm 0.109}$ & $\mathbf{0.184 \pm 0.08}$ & $\mathbf{0.087 \pm 0.042}$ \\ 
\bottomrule
\end{tabular}
}
\end{table*}

%% file: Tables/user_study.tex
\begin{table}[t]
\centering
\caption{\textbf{User Study Results} (mean ranking out of 4). Lower scores indicate better performance. Metrics include Temporal Coherence, Fidelity, Semantic Coherence, and Overall Score. All methods were evaluated using SD1.5. EIDT-V achieves the best overall ranking (1.9), followed closely by FreeBloom.}
\label{tab:userstudy}
\resizebox{\columnwidth}{!}{%
\begin{tabular}{lcccc}
\toprule
\textbf{Method} & \textbf{Temporal} & \textbf{Fidelity} & \textbf{Semantic} & \textbf{Overall} \\
 & \textbf{Coherence} ($\downarrow$) & ($\downarrow$) & \textbf{Coherence} ($\downarrow$) & ($\downarrow$) \\
\midrule
\textbf{FreeBloom} & 2.5 & \textbf{2.0} & \textbf{2.0} & 2.3 \\
\textbf{T2VZero} & 2.4 & 2.5 & 2.7 & 2.5 \\
\textbf{DirecT2V} & 3.2 & 3.5 & 3.1 & 3.3 \\
\textbf{EIDT-V} & \textbf{1.9} & \textbf{2.0} & 2.1 & \textbf{1.9} \\
\bottomrule
\end{tabular}%
}
\end{table}

%% file: Tables/ablation.tex
\begin{table}[t]
\centering
\caption{\textbf{Ablation Study Results} for MS-SSIM, LPIPS, and Temporal Consistency Loss across different configurations: ChatGPT (CG), Our Framewise Prompts (OFP), and each with Grid Prompt Switching (GrPS). The inclusion of GrPS, particularly with OFP, demonstrates substantial improvements in frame coherence, perceptual similarity, and temporal stability.}
\label{tab:ablation}
\resizebox{\columnwidth}{!}{%
\begin{tabular}{lccc}
\toprule
\textbf{Config.} & \textbf{MS-SSIM}~($\uparrow$) & \textbf{LPIPS}~($\downarrow$) & \textbf{Temp. Cons.}~($\downarrow$) \\
\midrule
CG & $0.132 \pm 0.052$ & $0.723 \pm 0.043$ & $0.389 \pm 0.046$ \\
OFP & $0.124 \pm 0.082$ & $0.686 \pm 0.062$ & $0.403 \pm 0.078$ \\
CG + GrPS & $0.588 \pm 0.129$ & $0.384 \pm 0.094$ & $0.157 \pm 0.062$ \\
OFP + GrPS & $\mathbf{0.63 \pm 0.137}$ & $\mathbf{0.33 \pm 0.1}$ & $\mathbf{0.152 \pm 0.062}$ \\

\bottomrule
\end{tabular}%
}
\end{table}

%% file: Figures/ablation_qual.tex
\begin{figure}[t]
\centering
\begin{tabular}{c}
    \includegraphics[width=0.8\columnwidth]{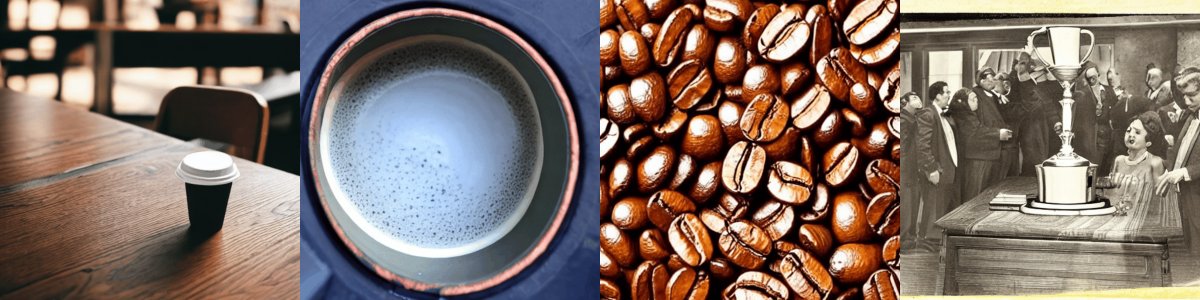} \\
    \scriptsize (a) CG without GrPS \\
    
    \includegraphics[width=0.8\columnwidth]{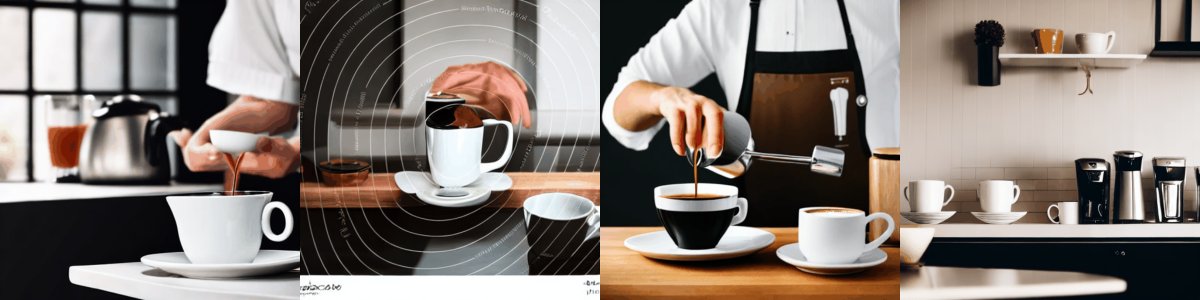} \\
    \scriptsize (b) OFP without GrPS \\

    \includegraphics[width=0.8\columnwidth]{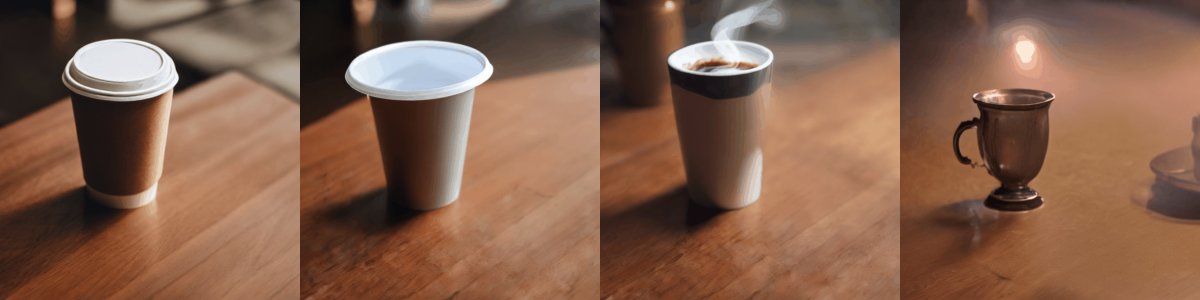} \\
    \scriptsize (c) CG with GrPS \\
    
    \includegraphics[width=0.8\columnwidth]{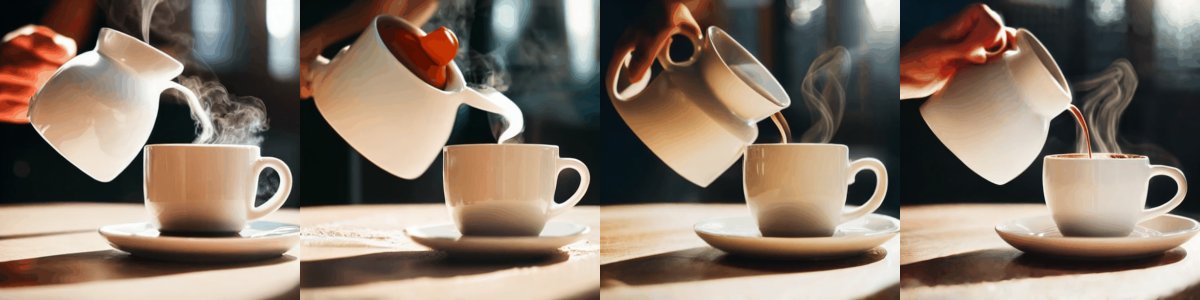} \\
    \scriptsize (d) OFP with GrPS \\
\end{tabular}
\caption{\textbf{Ablation Qualitative Results} (see \cref{tab:ablation}). Each row displays four equally spaced frames from the generated GIF. The prompt is ``\textbf{A cup of coffee being poured with steam rising}". The naive approach produces images linked only by theme. Applying OFP without GrPS offers minor improvements while incorporating GrPS (CG with GrPS) notably increases coherence. Finally, combining OFP with GrPS yields the best performance.}
\label{fig:ablation}
\end{figure}

%% file: sec/_6_Discussion.tex
\section{Discussion and Future Directions}
\label{sec:discussion_future}

Our approach achieves notable temporal coherence and visual fidelity across various architectures. By constraining the model in novel ways, we open a new path for training-free video generation that excels in producing subtle, targeted variations to improve frame-to-frame consistency.

A key aspect of our method is using variance as a proxy for motion. Since the model inherently understands only variance, we rely on text conditioning to direct this variance toward generating sequences that appear as coherent movements. This approach assumes that the text prompts will guide the model in applying variance in a way that visually represents a moving object. While this method proves effective in many scenarios, it has limitations. Occasionally, the model may generate frames with some visual inconsistencies, replacing expected movement with minimal changes that do not fully convey natural motion. 

Artifacts are also an issue. Distortions like limb elongation appear across training-free methods, partly due to strong conditioning effects, with some base models—most notably SD3—being especially prone. In our method, significant prompt changes late in the diffusion process may prevent full refinement of fine details. We evaluated 24 variants of a horse-running sequence, adjusting hyperparameters. Although no configuration eliminates artifacts, specific settings significantly reduce these distortions.

Future research could address these limitations by incorporating methods to improve frame coherence, such as minimal training to reinforce the distinction between variance and motion. A hybrid approach that combines targeted generation with light training, similar to techniques used in AnimateDiff \cite{guo_animatediff_2024}, could enable a low-cost, trained video generator with improved motion consistency. Expanding this approach within a trained, scalable environment could enhance adaptability, potentially leading to robust and resource-efficient tools for high-quality video generation.

%% file: sec/_7_Conclusions.tex
\section{Conclusion}
\label{sec:conclusion}

This paper presents a novel, training-free approach to video generation. We address critical challenges in achieving temporal consistency and architectural flexibility, leveraging core diffusion mechanisms and a grid-based prompt-switching strategy to produce coherent and realistic video sequences without requiring architectural modifications.

This work's primary contribution demonstrates that targeted variance, guided by text-based conditioning, can effectively substitute for more complex mechanisms in achieving visually coherent sequences. This approach has significant implications for enhancing the accessibility and scalability of video generation tools, narrowing the gap between high-quality output and low computational demands.

While the method has some limitations in differentiating targeted variance from actual motion, it lays a foundation for further exploration in resource-efficient video synthesis. Potential extensions include integrating lightweight training mechanisms or additional coherence-enhancing strategies to capture natural motion better and improve robustness. Overall, this study introduces a flexible, efficient, and high-fidelity video generation framework, offering valuable insights and tools for advancing the field of generative modeling.

%% file: sec/_8_Acknowledgments.tex
\section{Acknowledgments} 

This work was supported by the Engineering and Physical Sciences Research Council (EPSRC) under Grant No. EP/T518013/1 and Grant No. EP/Y021614/1. For the purpose of open access, the authors have applied a Creative Commons Attribution (CC BY) licence to any Author Accepted Manuscript version arising. The authors also acknowledge the University of Bath for providing access to the HEX high-performance computing (HPC) system, which was used for testing.

%% file: X_suppl.tex
\clearpage
\setcounter{page}{1}

\maketitlesupplementary

\input{Appendix/test_prompts}

\input{Appendix/hyperparam}

\input{Appendix/ip_adapter}

\input{Appendix/clip_results}

\input{Appendix/user_study}

\input{Appendix/additional_technical_details}

\input{Appendix/large_scale_changes}

% \onecolumn
\input{Appendix/additional_qualitative}
\input{Appendix/additional_best}
% \twocolumn

% ---
% \input{Appendix/original}

%% file: Appendix/test_prompts.tex
\section{Test Prompts}
\label{sup:test_prompts}

This section details the 50 prompts used to generate videos for our evaluation, along with the rationale behind their selection. The prompts were carefully designed to span a wide range of scenarios, including natural phenomena, object transformations, motion dynamics, and creative interpretations. This diversity ensures a comprehensive assessment of the model’s capabilities across various aspects of text-to-video generation.

\begin{enumerate}
    \item \textbf{A flower blooming from a bud to full bloom over time.}  
    \textit{Rationale:} Evaluates the model’s ability to depict time-lapse growth with smooth transitions.

    \item \textbf{A cat chasing a laser pointer dot across the room.}  
    \textit{Rationale:} Tests motion tracking and dynamic object interactions.

    \item \textbf{A rotating 3D cube changing colors.}  
    \textit{Rationale:} Assesses rendering of 3D rotation and color transitions.

    \item \textbf{A sunrise over the mountains turning into daytime.}  
    \textit{Rationale:} Evaluates depiction of natural phenomena and transitions in lighting conditions.

    \item \textbf{A person morphing into a wolf under a full moon.}  
    \textit{Rationale:} Challenges the model’s ability to handle complex transformations and creative scenarios.

    \item \textbf{Raindrops falling into a puddle creating ripples.}  
    \textit{Rationale:} Tests fluid dynamics rendering and subtle animation effects.

    \item \textbf{A city skyline transitioning from day to night with lights turning on.}  
    \textit{Rationale:} Evaluates handling of complex lighting transitions in urban scenes.

    \item \textbf{An apple falling from a tree and bouncing on the ground.}  
    \textit{Rationale:} Assesses motion physics and interactions with gravity.

    \item \textbf{A hand drawing a circle on a whiteboard.}  
    \textit{Rationale:} Tests precision in hand movements and sequential drawing actions.

    \item \textbf{An ice cube melting into water.}  
    \textit{Rationale:} Evaluates the depiction of state changes from solid to liquid.

    \item \textbf{A rocket launching into space and disappearing into the stars.}  
    \textit{Rationale:} Tests sequential events and scale changes in dynamic scenarios.

    \item \textbf{A chameleon changing colors on a branch.}  
    \textit{Rationale:} Challenges the model’s ability to handle color transitions and blending with surroundings.

    \item \textbf{A balloon inflating and then popping.}  
    \textit{Rationale:} Evaluates expansion dynamics and sudden transitions.

    \item \textbf{A paper airplane flying across a classroom.}  
    \textit{Rationale:} Tests object motion within a setting and interactions with the environment.

    \item \textbf{Clouds forming and then dissipating in the sky.}  
    \textit{Rationale:} Assesses rendering of natural elements and gradual changes.

    \item \textbf{A cup of coffee being poured with steam rising.}  
    \textit{Rationale:} Tests liquid dynamics and fine details like steam.

    \item \textbf{A clock’s hands moving fast-forward from noon to midnight.}  
    \textit{Rationale:} Evaluates representation of time passage and object motion.

    \item \textbf{A caterpillar transforming into a butterfly.}  
    \textit{Rationale:} Tests depiction of life cycles and metamorphosis.

    \item \textbf{A book opening and pages flipping.}  
    \textit{Rationale:} Evaluates detailed object movements and sequential actions.

    \item \textbf{A snowman melting under the sun.}  
    \textit{Rationale:} Tests weather effects and melting animations.

    \item \textbf{A traffic light cycling from red to green.}  
    \textit{Rationale:} Evaluates color changes and timing sequences.

    \item \textbf{A fish jumping out of water and diving back in.}  
    \textit{Rationale:} Tests motion through different mediums and splash effects.

    \item \textbf{An artist painting a canvas with a brush.}  
    \textit{Rationale:} Assesses fine motor actions and the process of creation.

    \item \textbf{A spinning globe showing continents passing by.}  
    \textit{Rationale:} Tests rotational motion and geographical accuracy.

    \item \textbf{Leaves falling from a tree in autumn.}  
    \textit{Rationale:} Evaluates natural motions and seasonal transitions.

    \item \textbf{A car transforming into a robot.}  
    \textit{Rationale:} Challenges the model with complex object transformations.

    \item \textbf{A candle burning down with the flame flickering.}  
    \textit{Rationale:} Tests gradual reduction and subtle lighting effects.

    \item \textbf{A soccer ball being kicked into a goal.}  
    \textit{Rationale:} Assesses motion, action sequences, and interactions.

    \item \textbf{A river flowing through a forest.}  
    \textit{Rationale:} Evaluates fluid motion and natural scenery rendering.

    \item \textbf{A rainbow appearing after rain.}  
    \textit{Rationale:} Tests depiction of weather transitions and color spectrum rendering.

    \item \textbf{An eclipse where the moon passes in front of the sun.}  
    \textit{Rationale:} Assesses celestial motion and lighting effects.

    \item \textbf{A horse galloping across a field.}  
    \textit{Rationale:} Tests animal motion and interaction with natural environments.

    \item \textbf{Popcorn popping in a microwave.}  
    \textit{Rationale:} Evaluates rapid, random movements and cooking processes.

    \item \textbf{A kaleidoscope pattern changing shapes and colors.}  
    \textit{Rationale:} Tests abstract patterns and continuous transformations.

    \item \textbf{A glass shattering into pieces when dropped.}  
    \textit{Rationale:} Challenges the model with sudden fragmentation and physics.

    \item \textbf{An astronaut floating in space waving.}  
    \textit{Rationale:} Tests human figures and movement in zero gravity.

    \item \textbf{A plant growing from a seed to a sapling.}  
    \textit{Rationale:} Evaluates depiction of growth over time.

    \item \textbf{Fireworks exploding in the night sky.}  
    \textit{Rationale:} Tests bright, dynamic visuals in a dark setting.

    \item \textbf{A dog wagging its tail happily.}  
    \textit{Rationale:} Assesses animal emotions and natural movements.

    \item \textbf{A compass needle spinning and settling north.}  
    \textit{Rationale:} Tests rotational motion and stabilization dynamics.

    \item \textbf{An umbrella opening up during rainfall.}  
    \textit{Rationale:} Evaluates object transformations and interactions with weather.

    \item \textbf{A stop-motion animation of clay figures moving.}  
    \textit{Rationale:} Tests frame-by-frame animation styles.

    \item \textbf{A battery draining from full to empty.}  
    \textit{Rationale:} Assesses gradual representation of depletion over time.

    \item \textbf{A puzzle being assembled piece by piece.}  
    \textit{Rationale:} Evaluates sequential object placement and completion.

    \item \textbf{A windmill’s blades turning in the breeze.}  
    \textit{Rationale:} Tests rotational motion influenced by wind.

    \item \textbf{A snake slithering through the grass.}  
    \textit{Rationale:} Assesses complex body movements in a natural setting.

    \item \textbf{A paintbrush changing colors as it moves.}  
    \textit{Rationale:} Tests motion-linked color transitions.

    \item \textbf{A volcano erupting with lava flowing.}  
    \textit{Rationale:} Evaluates dynamic natural events and fluid motion.

    \item \textbf{An eye blinking slowly.}  
    \textit{Rationale:} Tests subtle facial movements and precise timing.

    \item \textbf{A paper crumpling into a ball.}  
    \textit{Rationale:} Challenges the model with complex folding and texture changes.
\end{enumerate}

These prompts ensure a diverse evaluation of model capabilities, covering natural phenomena, motion dynamics, and creative transformations.

%% file: Appendix/hyperparam.tex
\section{Hyperparameter Selection}
\label{sup:hyperparameter_selection}

Hyperparameter tuning was a critical step in optimizing the performance of our video generation models, particularly with respect to temporal coherence, visual fidelity, and prompt adherence. We conducted a systematic grid search for SDXL and performed manual tuning for SD1.5 and SD3 to identify the most effective configurations for each model.

\subsection{Hyperparameters Considered}

The following key hyperparameters were explored during the grid search:

\begin{itemize}
    \item \textbf{Batch Size:} Values of 1, 2, and 3 were tested to balance GPU memory usage and frame coherence. Larger batch sizes can improve smoothness across frames by enabling better context preservation but increase memory requirements.
    
    \item \textbf{Intersection Strategy:} To ensure temporal continuity between frames, two strategies were compared:
    \begin{itemize}
        \item \textbf{First:} Each frame intersects with a static base image (batch size = num frames - 1).
        \item \textbf{Previous:} Each frame intersects with the last frame from the previous batch.
    \end{itemize}
    
    \item \textbf{Guidance Scale:} A range of values from 3.0 to 13.0 was tested to balance adherence to text prompts against visual diversity. Higher values generally emphasize prompt alignment but may reduce variability.

    \item \textbf{Multi-Prompt Strategy:} For models supporting multiple text inputs, we evaluated different strategies:
    \begin{itemize}
        \item \textbf{PreviousFrame:} Using the text of the previous frame as secondary input.
        \item \textbf{BaseFrame:} Using the text of the first frame as secondary input.
        \item \textbf{VideoText:} Using the user’s text input as secondary input throughout the sequence.
    \end{itemize}
    
    \item \textbf{Falloff:} This hyperparameter controls the degree of variability by raising attention mappings to a power. Higher falloff values reduce areas of variation, leading to greater temporal stability but potentially limiting variance.
\end{itemize}

\subsection{Grid Search Strategy}

The grid search was primarily conducted on the SDXL model, utilizing a diverse set of prompts and systematically varying hyperparameters. Each configuration was evaluated using the following metrics:
\begin{itemize}
    \item \textbf{Multi-Scale Structural Similarity (MS-SSIM):} Measures structural similarity between consecutive frames to evaluate content preservation.
    \item \textbf{Learned Perceptual Image Patch Similarity (LPIPS):} Analyzes perceptual similarity by comparing high-level features across frames.
    \item \textbf{Temporal Consistency Loss:} Assesses smoothness of motion using optical flow analysis.
\end{itemize}

For each configuration, these metrics were normalized to a [0, 1] range, and an equally weighted combined loss function was used for evaluation:
\begin{align}
\text{Combined Loss} &= (1 - \text{Normalized MS-SSIM}) \nonumber \\
&\quad + \text{Normalized LPIPS} \nonumber \\
&\quad + \text{Normalized Temporal Consistency Loss}
\end{align}

Lower combined loss values indicate better overall performance. We analyzed the most frequent high-performing hyperparameter configurations to identify optimal settings.

\subsection{Results and Empirical Best Settings}

From the grid search and manual tuning, the following configurations emerged as optimal for each model:

\subsubsection{SDXL}
\begin{itemize}
    \item \textbf{Batch Size:} 3
    \item \textbf{Intersection Strategy:} Previous
    \item \textbf{Multi-Prompt Strategy:} VideoText
    \item \textbf{Guidance Scale:} 11.0
    \item \textbf{Falloff:} 2
\end{itemize}

\subsubsection{SD1.5}
\begin{itemize}
    \item \textbf{Batch Size:} 3
    \item \textbf{Intersection Strategy:} Previous
    \item \textbf{Guidance Scale:} 11.0
    \item \textbf{Falloff:} 2
\end{itemize}

\subsubsection{SD3}
Manual testing revealed the following optimal settings for SD3:
\begin{itemize}
    \item \textbf{Batch Size:} 2
    \item \textbf{Intersection Strategy:} Previous
    \item \textbf{Multi-Prompt Strategy:} VideoText / none
    \item \textbf{Guidance Scale:} 9.0 / 11.0
    \item \textbf{Falloff:} 1
\end{itemize}

\subsection{Discussion}

The consistency of effective hyperparameters across models highlights general principles for optimizing video generation in diffusion-based models:
\begin{itemize}
    \item A \textbf{batch size of 3} achieves a balance between computational efficiency and temporal coherence.
    \item Using the \textbf{“Previous” intersection strategy} significantly enhances frame-to-frame continuity, reducing flickering and visual artifacts.
    \item A \textbf{guidance scale of 11.0} strikes an effective balance between adherence to text prompts and visual creativity.
    \item The \textbf{VideoText multi-prompt strategy} dynamically guides generation using the original text input and improves temporal consistency for supported architectures.
    \item \textbf{Falloff:} A falloff of 2 is ideal for SDXL and SD1.5, producing stable yet diverse outputs, whereas a falloff of 1 is better suited for SD3, maintaining sufficient variability.
\end{itemize}

These findings provide a robust framework for optimizing diffusion models for video generation tasks and offer a foundation for further experimentation and refinement.

%% file: Appendix/ip_adapter.tex
\section{IP-Adapter}
\label{sup:ip_adapter}

In this section, we discuss the rationale for testing the IP-Adapter within our framework and evaluate its impact on video generation quality.

\subsection{Rationale for Using IP-Adapter}

The IP-Adapter \cite{ye_ip-adapter_2023} was integrated into our pipeline to leverage its cross-attention mechanism, which aligns with our modular and conditional generation objectives. As a well-established method in conditional image generation, the IP-Adapter provides fine-grained control over generated content by incorporating auxiliary inputs through attention mechanisms. This modular approach is more accessible than the architectural changes made by previous works in this area.

\subsection{Results with IP-Adapter}

The performance impact of the IP-Adapter is summarized in \cref{tab:quantitative}. Key observations include:

\begin{itemize}
    \item \textbf{LPIPS:} A slight improvement was observed, with scores improving from $0.33 \pm 0.1$ (without IP-Adapter) to $0.316 \pm 0.089$ (with IP-Adapter). This suggests a marginal enhancement in perceptual quality.
    \item \textbf{MS-SSIM:} A modest increase in structural similarity was noted, with scores rising from $0.63 \pm 0.137$ (without IP-Adapter) to $0.655 \pm 0.13$ (with IP-Adapter).
    \item \textbf{Temporal Consistency Loss:} Negligible changes were observed, indicating that the IP-Adapter had limited impact on improving frame-to-frame coherence.
\end{itemize}

While these results highlight minor improvements in perceptual quality and structural similarity, the observed gains fall within the standard deviation, raising questions about their statistical significance.

\subsection{Discussion on Results}

Although the IP-Adapter provided minor enhancements in certain metrics, the improvements were not substantial enough to justify the added complexity it introduces into the pipeline. Given the lack of significant impact on temporal consistency and the marginal nature of the improvements, we conclude that the IP-Adapter may not be well-suited for our specific zero-shot video generation framework.

%% file: Appendix/clip_results.tex
\section{CLIP Results}
\label{sup:clip_results}

\input{Tables/clip}

In this section, we present the CLIP scores for all models used in our main qualitative experiments. The results, summarized in \cref{tab:clip_score_comparison}, reveal minimal variance in CLIP scores across different models and configurations. While CLIP scores effectively measure text-image alignment, they do not correlate strongly with video generation performance or quality.

\subsection{Analysis of CLIP Scores}

As shown in \cref{tab:clip_score_comparison}, the CLIP scores for all models and configurations have vary little variance between them. Key observations include:

\begin{itemize}
    \item \textbf{SD1.5-based models:} Scores ranged from $0.271 \pm 0.022$ (Free-Bloom) to $0.294 \pm 0.026$ (T2V-Zero). Our proposed method achieved scores of $0.278 \pm 0.03$ and $0.271 \pm 0.034$ across different configurations.
    
    \item \textbf{Newer models:} Both SD3 and SDXL achieved comparable scores, with $0.276 \pm 0.028$ and $0.271 \pm 0.031$, respectively.
\end{itemize}

Noting that the video output of these models was significantly different, these results demonstrate that while CLIP scores effectively fail to capture essential aspects of video quality, such as temporal coherence and perceptual fidelity. To address this we propose the three metrics we use. Details can be found in the main text.

%% file: Tables/clip.tex
\begin{table}[H]
\centering
\caption{Quantitative comparison of CLIP Score for our method and previous works.}
\label{tab:clip_score_comparison}
\resizebox{\linewidth}{!}{%
\begin{tabular}{ll|c}
\toprule
\textbf{Method} & \textbf{Pre-Trained Model} & \textbf{CLIP Score} \\ 
\midrule
\textbf{DirecT2V} & \textbf{SD1.5} & $0.276 \pm 0.025$ \\ 
\textbf{Free-Bloom} & \textbf{SD1.5} & $0.271 \pm 0.022$ \\ 
\textbf{T2V-Zero} & \textbf{SD1.5} & $0.294 \pm 0.026$ \\ 
\textbf{Ours} & \textbf{SD1.5} & $0.278 \pm 0.03$ \\ 
\textbf{Ours} & \textbf{SD1.5 w/IP-Adapter\cite{ye_ip-adapter_2023}} & $0.271 \pm 0.034$ \\ 
\midrule
\textbf{Ours} & \textbf{SD3} & $0.276 \pm 0.028$ \\ 
\textbf{Ours} & \textbf{SDXL} & $0.271 \pm 0.031$ \\ 
\bottomrule
\end{tabular}
}
\end{table}

%% file: Appendix/user_study.tex
\section{User Study Setup}
\label{sup:user_study_setup}

This section details the setup and execution of the user study conducted to validate the comparative performance of our video generation models.

\subsection{Study Design}

The user study was designed to evaluate the performance of our SD1.5 model against other SD1.5-based baseline models using 50 video prompts. Participants were asked to assess the generated videos across four evaluation criteria:

\begin{enumerate}
    \item \textbf{Smoothness (Temporal Coherence):} The quality of transitions between frames, avoiding jumps or awkward motion.
    \item \textbf{Picture Quality (Fidelity):} The visual fidelity and clarity of the video frames.
    \item \textbf{Adherence to Description (Semantic Coherence):} How accurately the video aligned with the given text prompt.
    \item \textbf{Overall Quality:} A holistic evaluation incorporating all three criteria.
\end{enumerate}

Each video prompt was presented as four GIFs, corresponding to outputs from different models. The GIFs were randomly assigned labels (A, B, C, D) to eliminate potential biases. Participants ranked the GIFs for each evaluation criterion in descending order of preference (e.g., if A is preferred ABCD).

\subsection{Study Implementation}

The study was implemented as an interactive web application, allowing participants to evaluate videos in a structured and intuitive manner. The code for this web app will also be made public with the rest of the code. Key features of the study setup included:

\begin{itemize}
    \item \textbf{Randomized Presentation:} GIFs for each video prompt were shuffled and assigned randomized labels for each participant.
    \item \textbf{Ranking Interface:} A simple ranking system required participants to assign a unique rank (1 to 4) to each GIF for all four criteria.
    \item \textbf{Data Collection:} Responses were validated to ensure completeness (e.g., each letter A, B, C, and D appeared exactly once per ranking) and stored in CSV format for aggregation and analysis.
\end{itemize}

Clear instructions were provided to ensure participants understood the evaluation process and the significance of each criterion.

\subsection{Participant Details}

A total of eight participants were involved in the study. Each participant evaluated all 50 video prompts across the four criteria, resulting in a total of 1,600 individual rankings. Participants represented a mix of technical and non-technical backgrounds, from ages 17 to 55, ensuring a balanced perspective on video quality.

\subsection{Analysis and Observations}

The rankings were aggregated across participants to derive average scores and identify trends. Key observations included:

\begin{itemize}
    \item \textbf{High Variability in Preferences:} Standard deviations across rankings were consistently around 1 for all evaluation criteria, highlighting subjective variability in participant preferences.
    \item \textbf{Aggregated Insights:} Despite individual differences, the aggregated results consistently favored our model in terms of smoothness, picture quality, and adherence to descriptions.
\end{itemize}

Given the observed variability, we focused on aggregated rankings and qualitative trends rather than standard deviation as a primary metric.

\subsection{Conclusion}

The user study highlighted the strengths of our SD1.5 model in generating videos with superior smoothness, picture quality, and adherence to prompts compared to baseline models. While the small participant pool and the subjective nature of rankings introduced variability, the overall trends were consistent. Future studies involving a larger and more diverse participant base could further validate and refine these findings.

%% file: Appendix/additional_technical_details.tex
\section{Additional Technical Details}
\label{sup:additional_technical_details}

We used an 8B LLaMA \cite{dubey_llama_2024} model locally for prompt generation due to its practicality, but we also tested Qwen 2.5 7B \cite{qwen2} and Mistral 7B \cite{jiang2023mistral7b} (see \cref{tab:additional_quantitative}). As our model is designed to be LLM-agonistic, there were no significant differences in performance. Naturally, the in-context information may need to be optimized for each model, but in general, the LLaMa model performed best, which is why we used it in our main testing.

Our model also does not depend on any particular ODE solver; as such, we used the standard options provided in the Diffusers Library \cite{von-platen-etal-2022-diffusers}. 

We do not fix the seeds across models, as their internal sampling mechanisms can yield differing outputs even with a fixed seed. \cref{fig:horse_forward_comparison} demonstrates that distinct methods can produce substantially different results despite fixed seeds (and the same image generator). 

\cref{fig:compact_attention_examples} provides a detailed example of how the attention mechanism works. It shows an example of the different text components and how they are combined with a CLIP model to generate an attention map over the previous frame. This attention map highlights areas that require high variance. This allows the image generator to make more changes in the given region, and as we can see, the balloon has changed in the next frame.

\input{Figures/rebuttal_visualise}

\input{Tables/rebuttal_main}

%% file: Figures/rebuttal_visualise.tex
\begin{figure}[ht]
    \centering
    
    %%%%%%%%%%%%%%%%%%%%%%%%%%%%%%%%%%%%%%%%%%%%%%%%%%%%%%%%%%%%%
    %                     FIRST EXAMPLE                         %
    %%%%%%%%%%%%%%%%%%%%%%%%%%%%%%%%%%%%%%%%%%%%%%%%%%%%%%%%%%%%%
    \begin{minipage}[t]{0.3\linewidth}
        \centering
        \includegraphics[width=\linewidth]{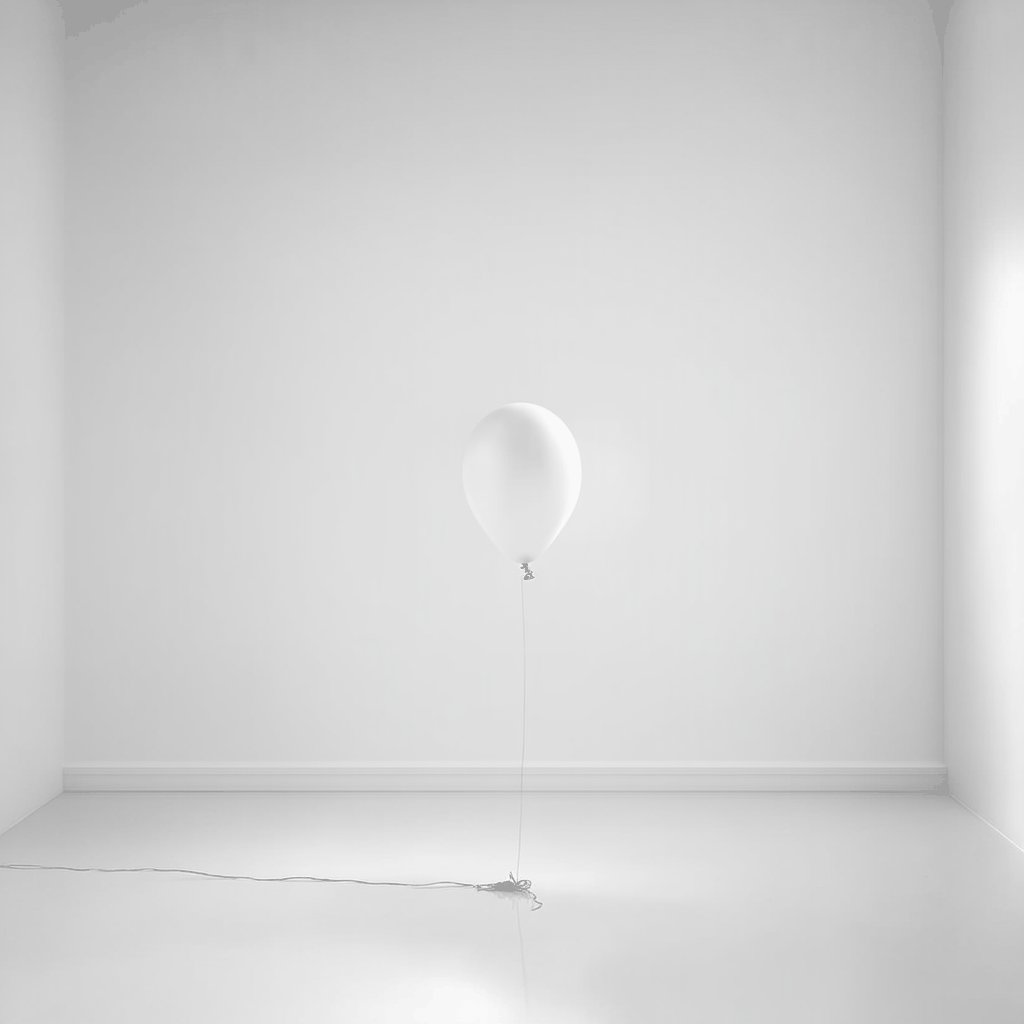}
        \caption*{\scriptsize Before}
    \end{minipage}
    \hfill
    \begin{minipage}[t]{0.3\linewidth}
        \centering
        \includegraphics[width=\linewidth]{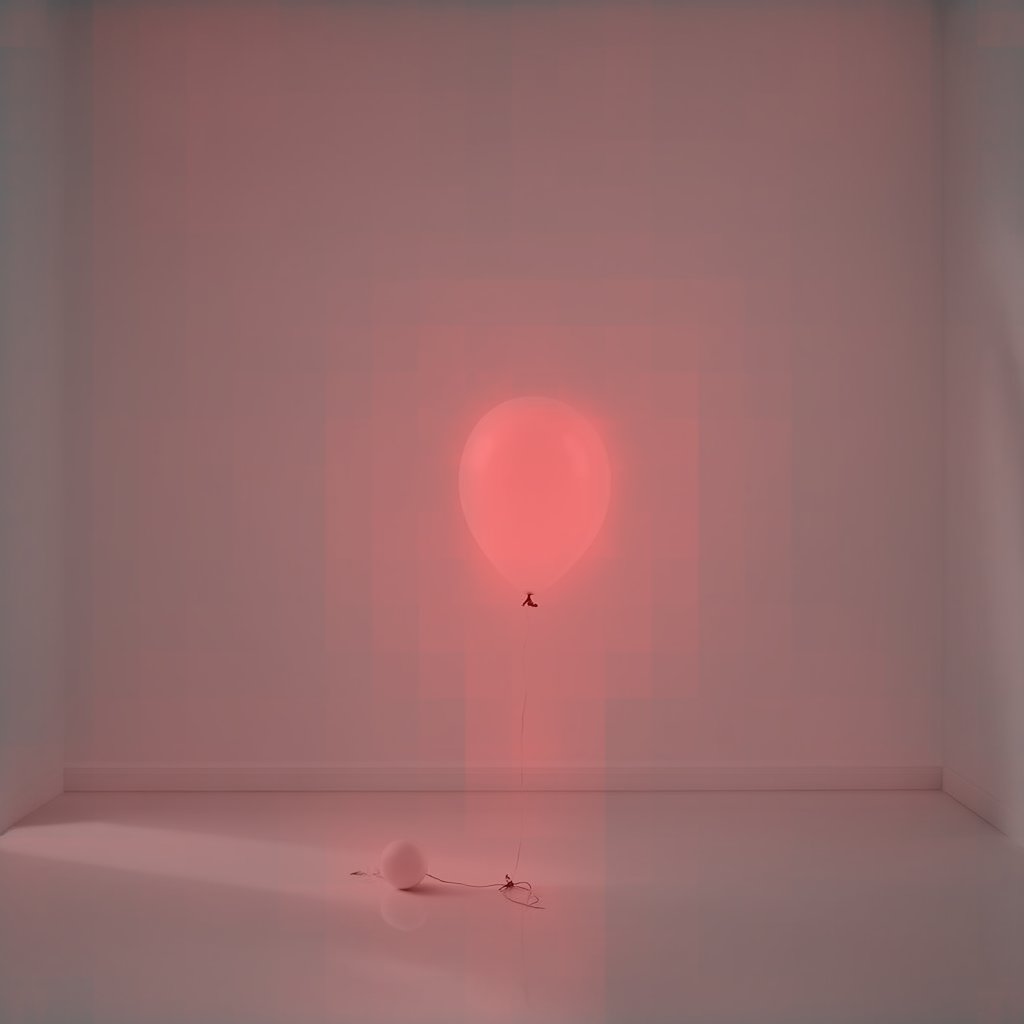}
        \caption*{\scriptsize Attention Map}
    \end{minipage}
    \hfill
    \begin{minipage}[t]{0.3\linewidth}
        \centering
        \includegraphics[width=\linewidth]{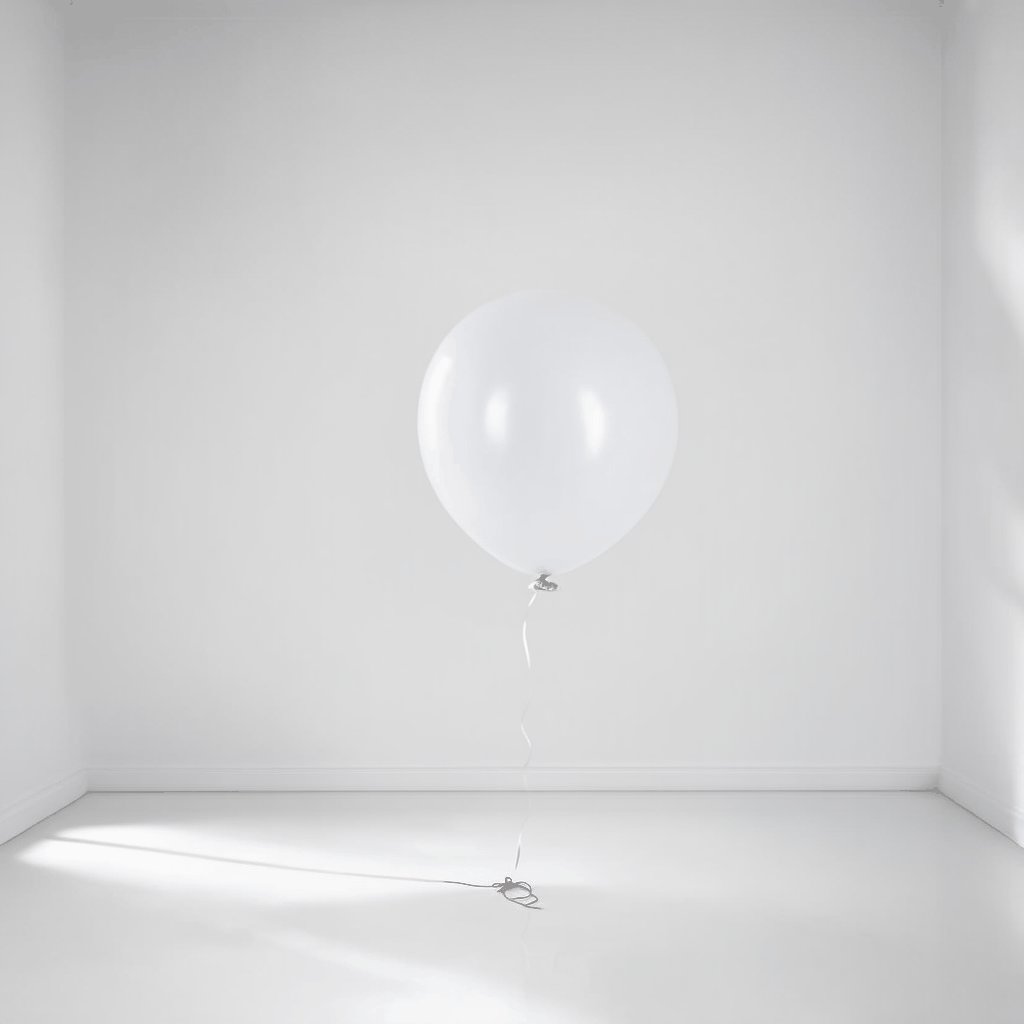}
        \caption*{\scriptsize Next Frame}
    \end{minipage}
    
    \scriptsize
    \textbf{Fixed part}: 
    \textit{Scene: Empty white background; a balloon tied to a string. Character: None. Setting: Indoor location; dim lighting; no distractions. Background: Simple, plain surface; minimal reflections.} \\
    \textbf{Previous Dynamic part}: \textit{Inflation begins slow.} \\
    \textbf{New Dynamic part}: \textit{Volume increases gradually.} \\
    \textbf{Differences Detected}: [“Balloon inflating”, “Volume increasing”]

    \caption{
    Our method detects differences, generates attention map and combines them by taking the maximal value at each pixel. 
    Bright red regions in attention correspond to high variance.}
    \label{fig:compact_attention_examples}
\end{figure}

%% file: Tables/rebuttal_main.tex
\begin{table}[h]
\centering
    \caption{Quantitative performance of EIDT-V using alternative LLMs. For more details, please refer \cref{tab:quantitative}.}
\label{tab:additional_quantitative}
\resizebox{\linewidth}{!}{%
\begin{tabular}{ll|c|ccc}
\toprule
\textbf{Method} & \textbf{Pre-Trained} & \textbf{Unmodified} & \textbf{MS-SSIM} & \textbf{LPIPS} & \textbf{Temporal} \\ 
 & \textbf{Model} & \textbf{Architecture} & ($\uparrow$) & ($\downarrow$) & \textbf{Consistency} ($\downarrow$) \\ 
\midrule

\textbf{EIDT-V} & \textbf{SD1.5 w/ Qwen LLM}      & \cmark & 0.572 ± 0.151 & 0.370 ± 0.093 & 0.168 ± 0.058 \\
\textbf{EIDT-V} & \textbf{SD1.5 w/ Mistral LLM}   & \cmark & 0.599 ± 0.122 & 0.353 ± 0.077 & 0.162 ± 0.061 \\
\bottomrule
\end{tabular}
}
\end{table}

%% file: Appendix/large_scale_changes.tex
\section{Large Scale Changes}
\label{sup:large_scale_changes}

Extreme scene changes (e.g., when the subject moves forward while the background moves in the opposite direction) are challenging for all training-free approaches. As shown in \cref{fig:horse_forward_comparison}, methods such as T2VZero and DirecT2V often fail to preserve the subject adequately, while FreeBloom exhibits excessive variation. In contrast, our method localizes changes, effectively balancing consistency and variance.

\input{Figures/rebuttal_horse_forward}

%% file: Figures/rebuttal_horse_forward.tex
\begin{figure}[h]
    \centering
    
    \footnotesize T2VZero\\
    \includegraphics[width=0.8\columnwidth]{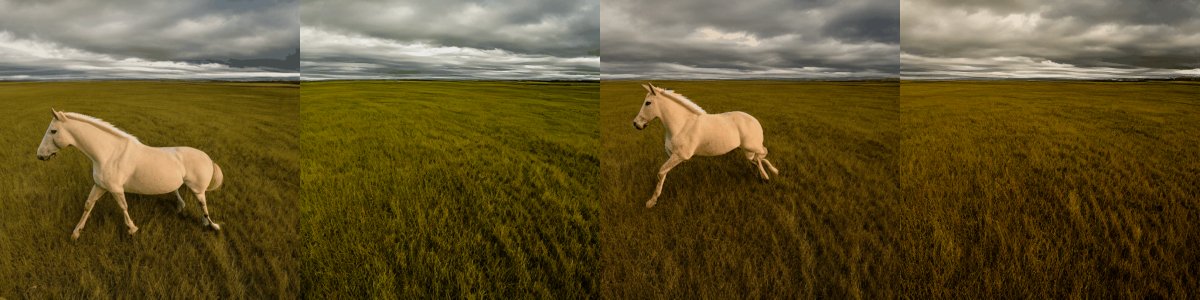}
    
    \footnotesize DirecT2V\\
    \includegraphics[width=0.8\columnwidth]{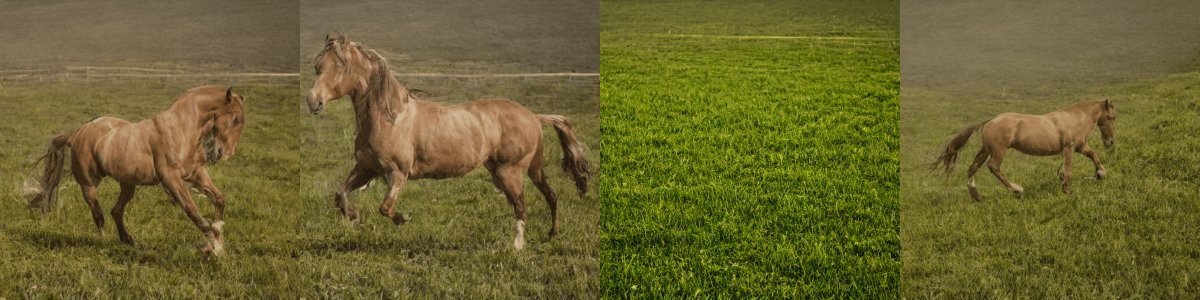}
    
    \footnotesize FreeBloom\\
    \includegraphics[width=0.8\columnwidth]{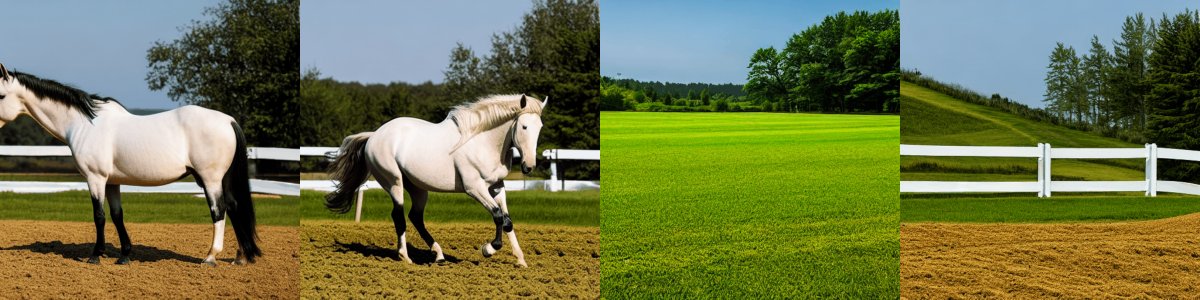}
    
    \footnotesize EIDT-V SD\\
    \includegraphics[width=0.8\columnwidth]{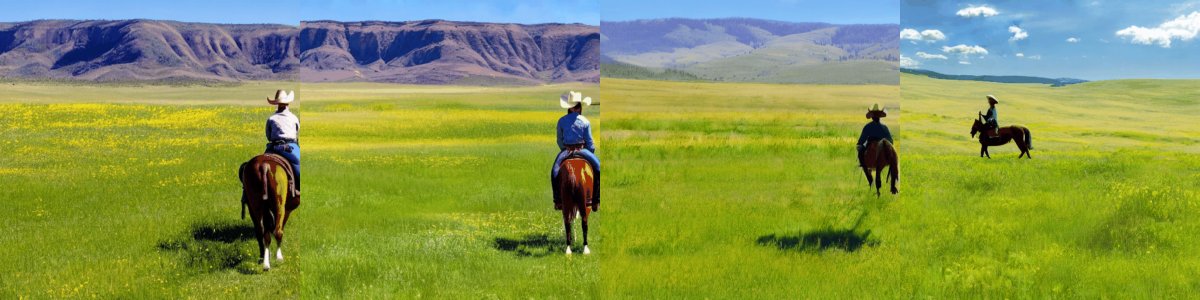}

    \caption{Qualitative comparison SD1.5 based video-generation models for the prompt: 
    ``\textbf{A first-person view from atop a horse, its ears and mane visible, moving forward across a grassy field}”. A fixed seed was used across all models.}
    \label{fig:horse_forward_comparison}
\end{figure}

%% file: Appendix/additional_qualitative.tex
\section{Additional Qualitative}
\label{sup:additional_qualitative}
\onecolumn 
\begin{figure}[H]
    \centering
    \begin{minipage}[t]{\textwidth}
        \centering
        \textbf{T2VZero} \\        \includegraphics[width=0.85\linewidth]{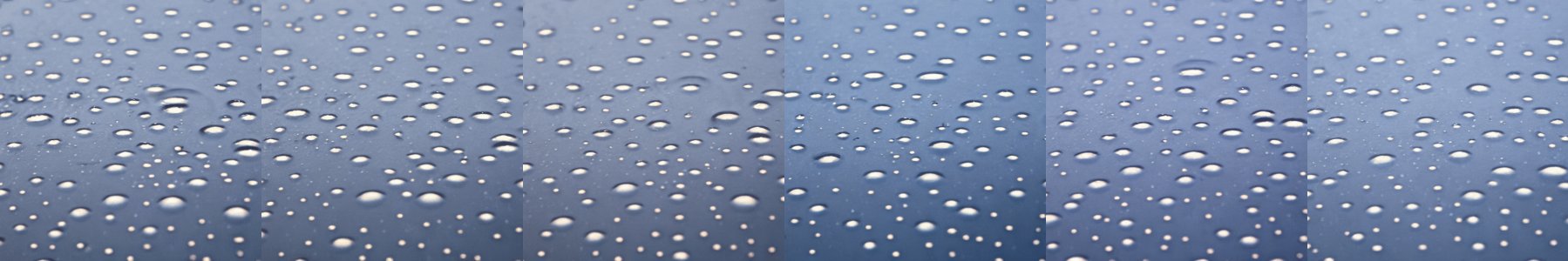}
    \end{minipage}
    
    \vspace{1em}
    \begin{minipage}[t]{\textwidth}
        \centering
        \textbf{DirecT2V} \\        \includegraphics[width=0.85\linewidth]{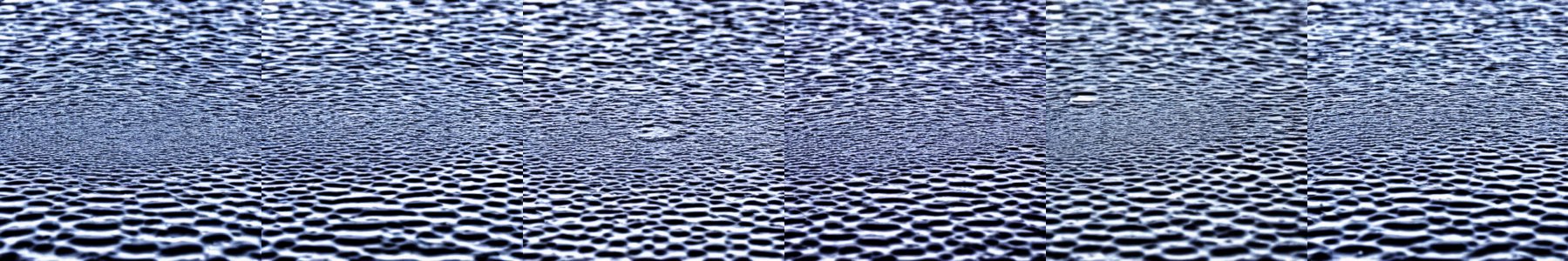}
    \end{minipage}
    
    \vspace{1em}
    \begin{minipage}[t]{\textwidth}
        \centering
        \textbf{FreeBloom} \\        \includegraphics[width=0.85\linewidth]{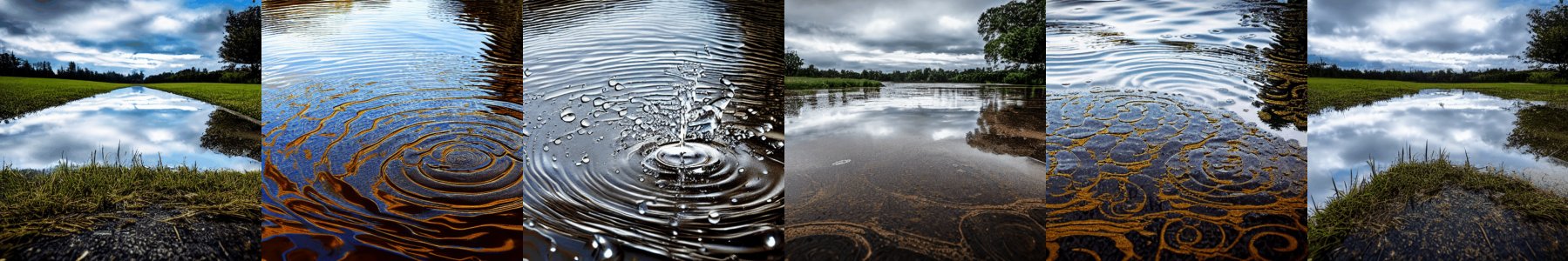}
    \end{minipage}
    
    \vspace{1em}
    \begin{minipage}[t]{\textwidth}
        \centering
        \textbf{EIDT-V SD} \\        \includegraphics[width=0.85\linewidth]{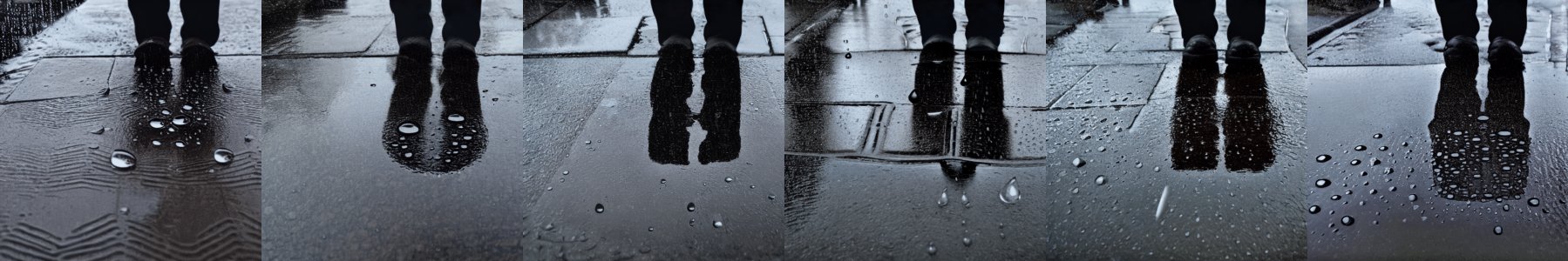}
    \end{minipage}
    
    \vspace{1em}
    \begin{minipage}[t]{\textwidth}
        \centering
        \textbf{EIDT-V SD\_IP} \\        \includegraphics[width=0.85\linewidth]{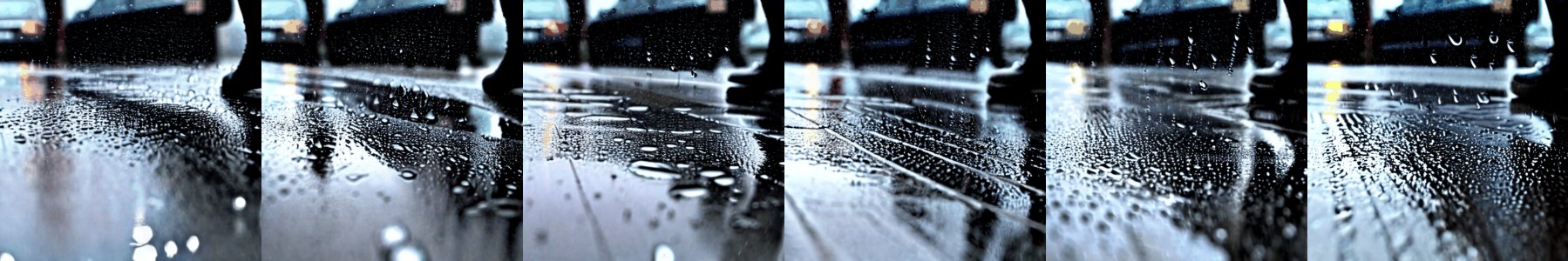}
    \end{minipage}
    
    \vspace{1em}
    \begin{minipage}[t]{\textwidth}
        \centering
        \textbf{EIDT-V SDXL} \\        \includegraphics[width=0.85\linewidth]{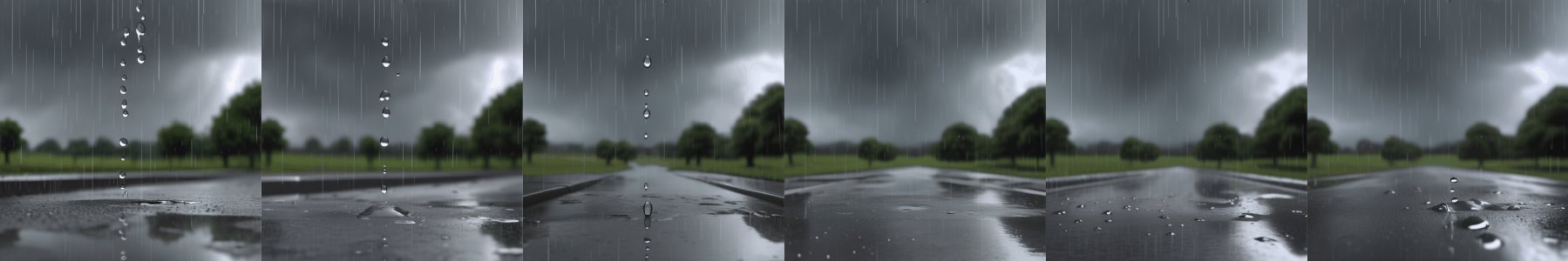}
    \end{minipage}
    
    \vspace{1em}
    \begin{minipage}[t]{\textwidth}
        \centering
        \textbf{EIDT-V SD3} \\        \includegraphics[width=0.85\linewidth]{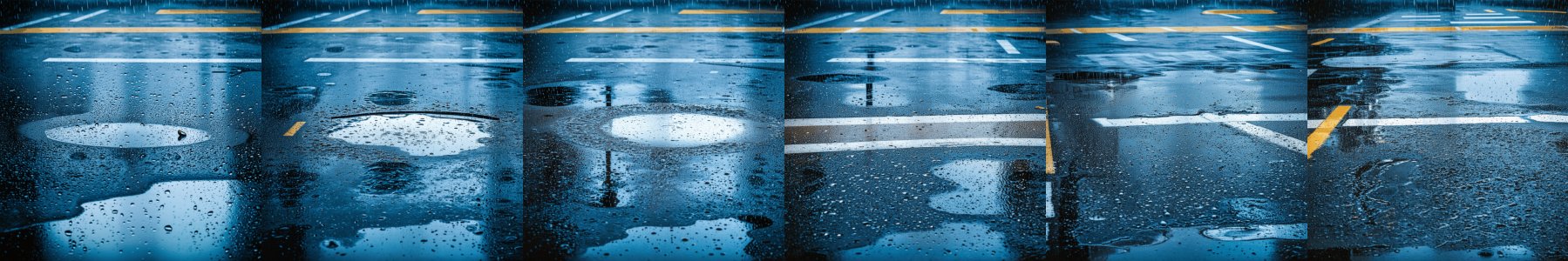}
    \end{minipage}
    
    \caption{Raindrops falling into a puddle creating ripples.}
\end{figure}
\twocolumn 
\onecolumn 
\begin{figure}[H]
    \centering
    \begin{minipage}[t]{\textwidth}
        \centering
        \textbf{T2VZero} \\
        \includegraphics[width=0.85\linewidth]{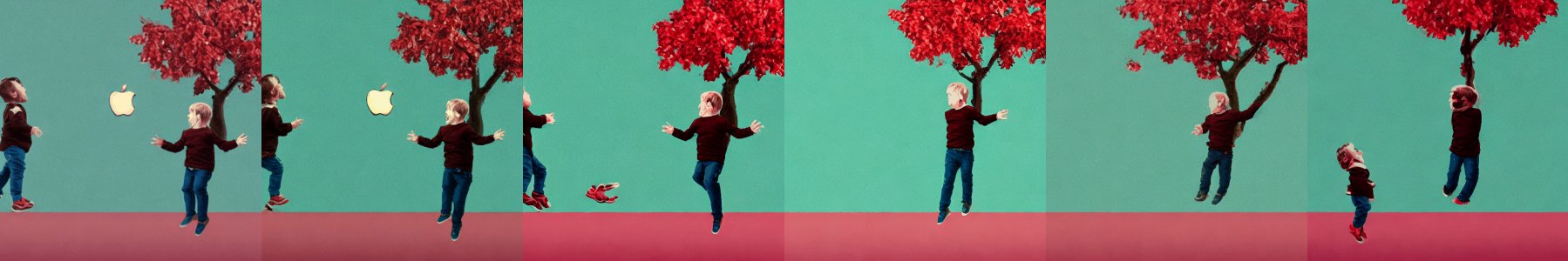}
    \end{minipage}

    \vspace{1em}

    \begin{minipage}[t]{\textwidth}
        \centering
        \textbf{DirecT2V} \\
        \includegraphics[width=0.85\linewidth]{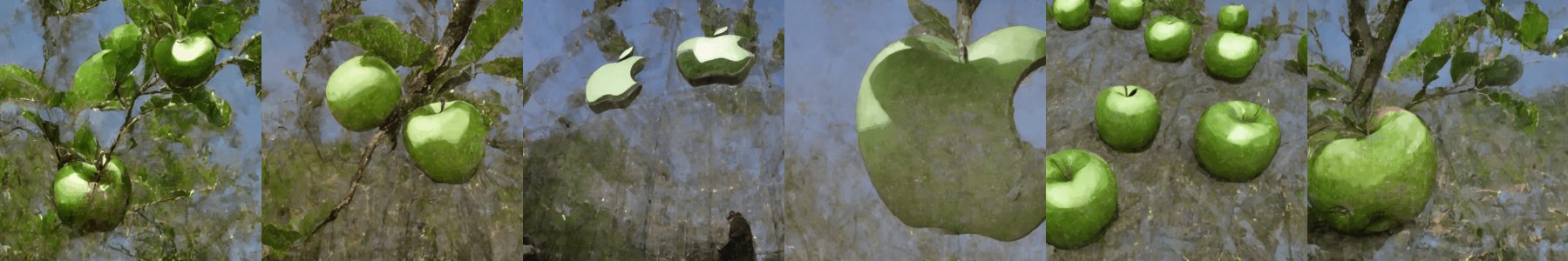}
    \end{minipage}

    \vspace{1em}

    \begin{minipage}[t]{\textwidth}
        \centering
        \textbf{FreeBloom} \\
        \includegraphics[width=0.85\linewidth]{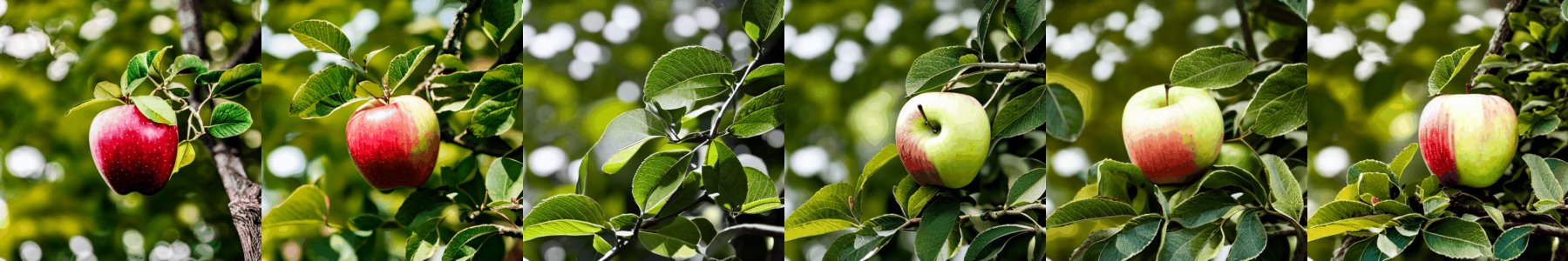}
    \end{minipage}

    \vspace{1em}

    \begin{minipage}[t]{\textwidth}
        \centering
        \textbf{EIDT-V SD} \\
        \includegraphics[width=0.85\linewidth]{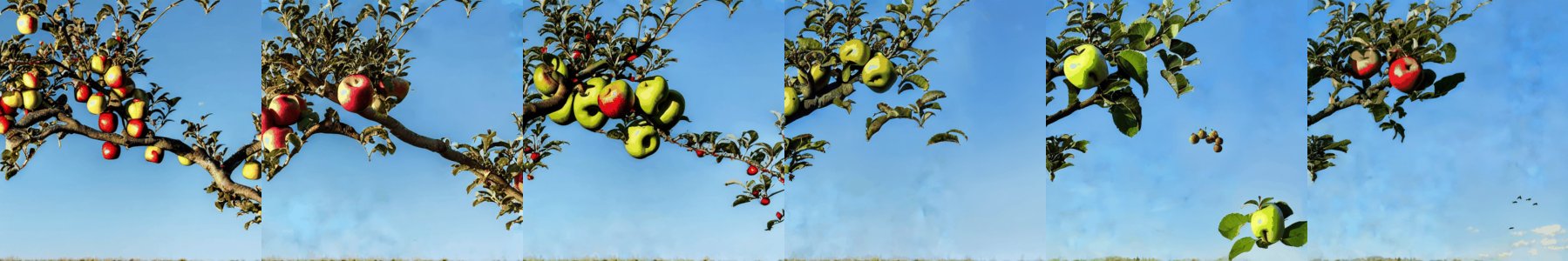}
    \end{minipage}

    \vspace{1em}

    \begin{minipage}[t]{\textwidth}
        \centering
        \textbf{EIDT-V SD\_IP} \\
        \includegraphics[width=0.85\linewidth]{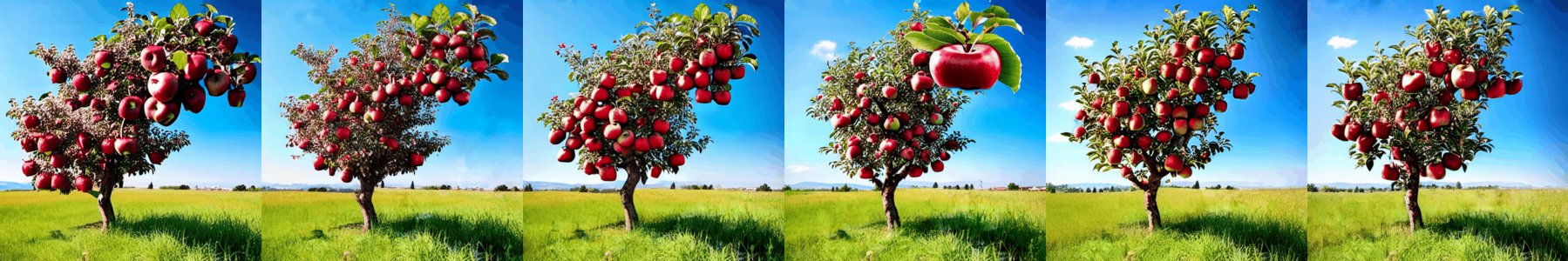}
    \end{minipage}

    \vspace{1em}

    \begin{minipage}[t]{\textwidth}
        \centering
        \textbf{EIDT-V SDXL} \\
        \includegraphics[width=0.85\linewidth]{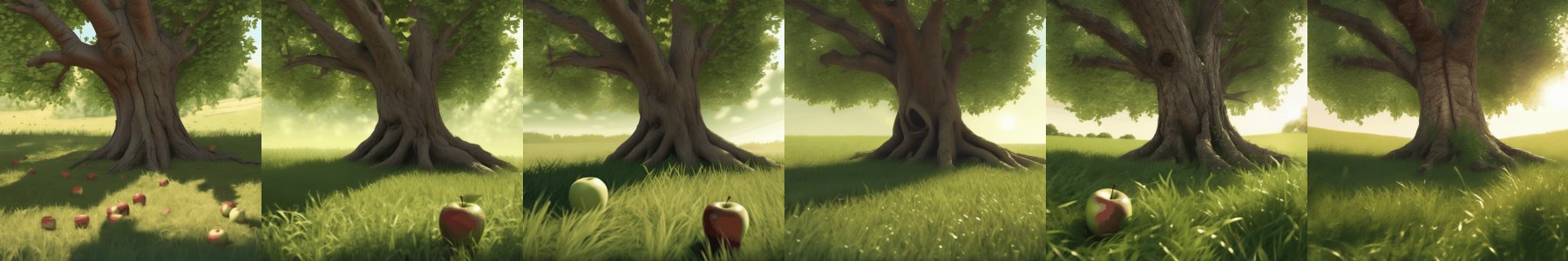}
    \end{minipage}

    \vspace{1em}

    \begin{minipage}[t]{\textwidth}
        \centering
        \textbf{EIDT-V SD3} \\
        \includegraphics[width=0.85\linewidth]{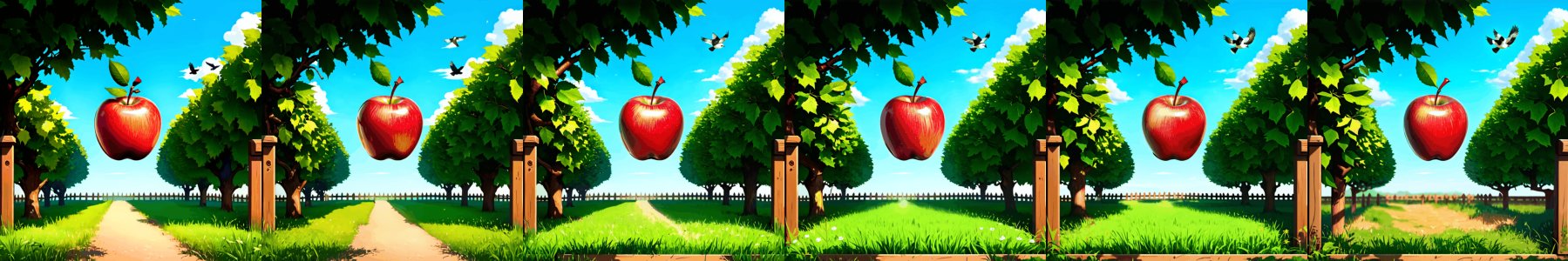}
    \end{minipage}

    \caption{An apple falling from a tree and bouncing on the ground.}
\end{figure}

\begin{figure}[H]
    \centering
    \begin{minipage}[t]{\textwidth}
        \centering
        \textbf{T2VZero} \\
        \includegraphics[width=0.85\linewidth]{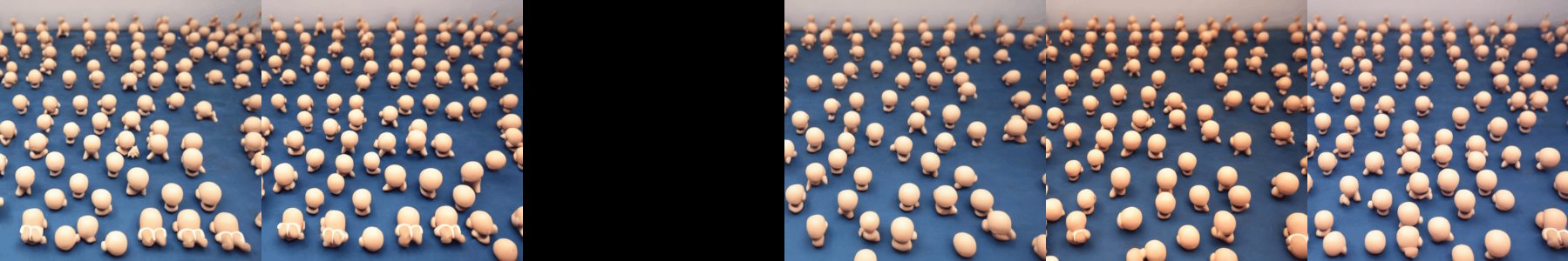}
    \end{minipage}

    \vspace{1em}

    \begin{minipage}[t]{\textwidth}
        \centering
        \textbf{DirecT2V} \\
        \includegraphics[width=0.85\linewidth]{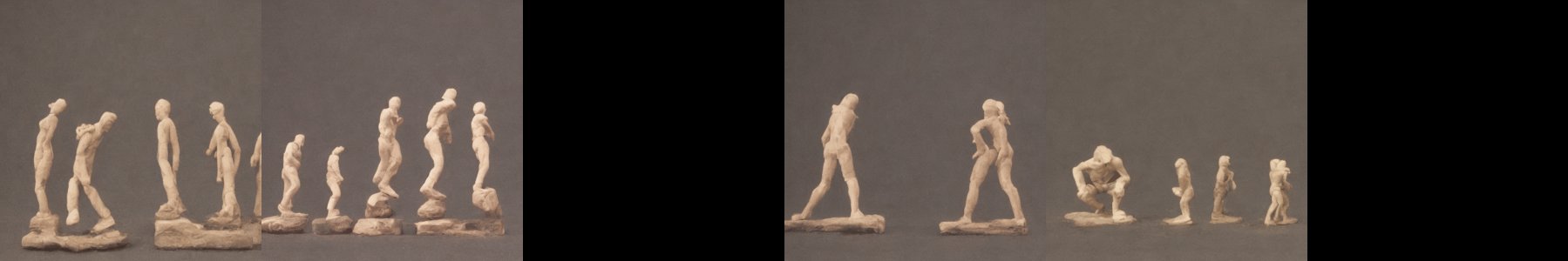}
    \end{minipage}

    \vspace{1em}

    \begin{minipage}[t]{\textwidth}
        \centering
        \textbf{FreeBloom} \\
        \includegraphics[width=0.85\linewidth]{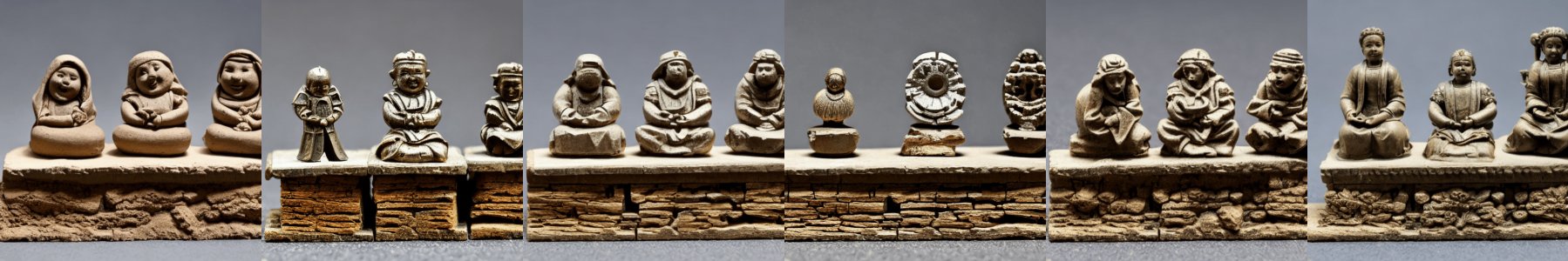}
    \end{minipage}

    \vspace{1em}

    \begin{minipage}[t]{\textwidth}
        \centering
        \textbf{EIDT-V SD} \\
        \includegraphics[width=0.85\linewidth]{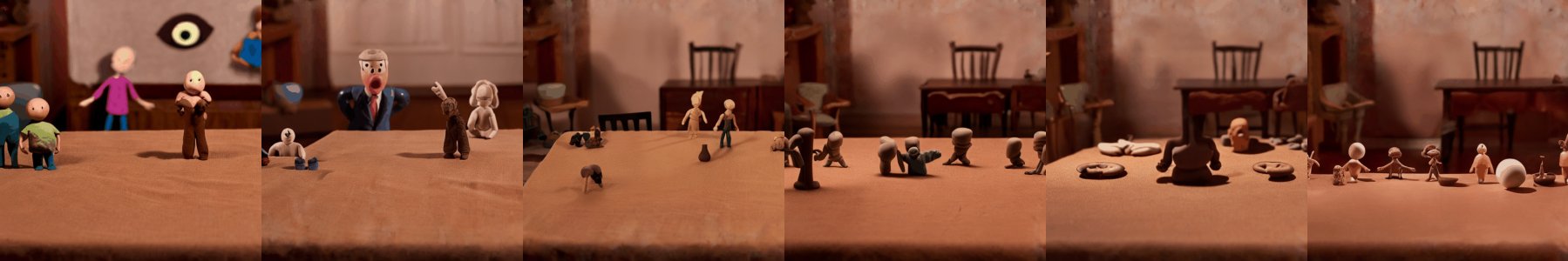}
    \end{minipage}

    \vspace{1em}

    \begin{minipage}[t]{\textwidth}
        \centering
        \textbf{EIDT-V SD\_IP} \\
        \includegraphics[width=0.85\linewidth]{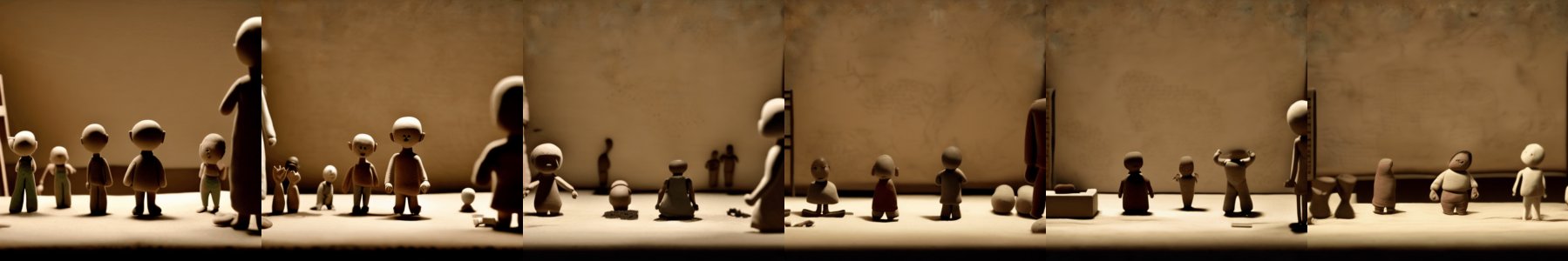}
    \end{minipage}

    \vspace{1em}

    \begin{minipage}[t]{\textwidth}
        \centering
        \textbf{EIDT-V SDXL} \\
        \includegraphics[width=0.85\linewidth]{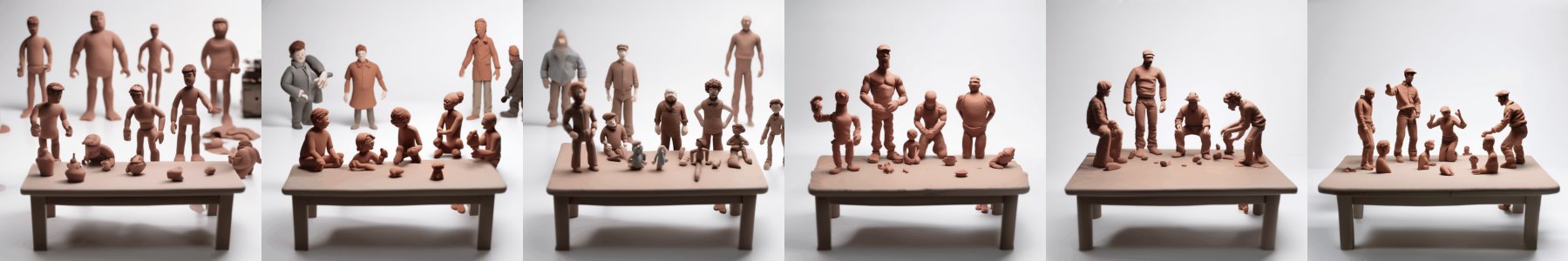}
    \end{minipage}

    \vspace{1em}

    \begin{minipage}[t]{\textwidth}
        \centering
        \textbf{EIDT-V SD3} \\
        \includegraphics[width=0.85\linewidth]{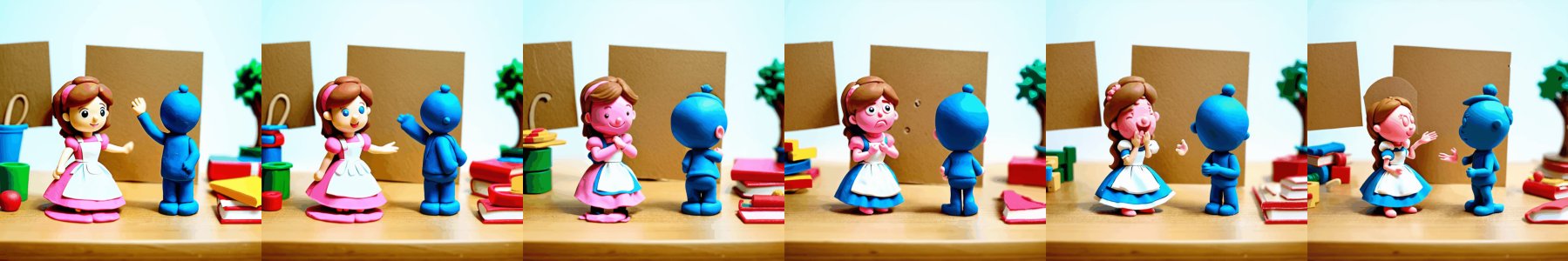}
    \end{minipage}

    \caption{A stop-motion animation of clay figures moving.}
\end{figure}

\begin{figure}[H]
    \centering
    \begin{minipage}[t]{\textwidth}
        \centering
        \textbf{T2VZero} \\
        \includegraphics[width=0.85\linewidth]{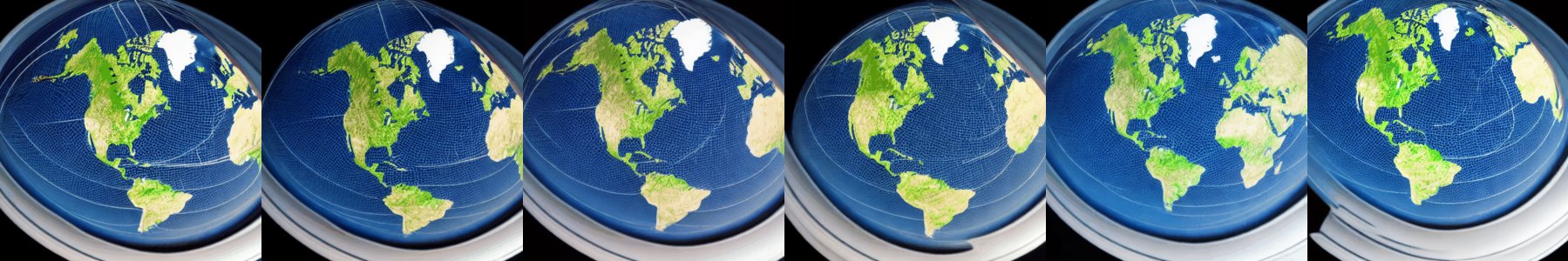}
    \end{minipage}

    \vspace{1em}

    \begin{minipage}[t]{\textwidth}
        \centering
        \textbf{DirecT2V} \\
        \includegraphics[width=0.85\linewidth]{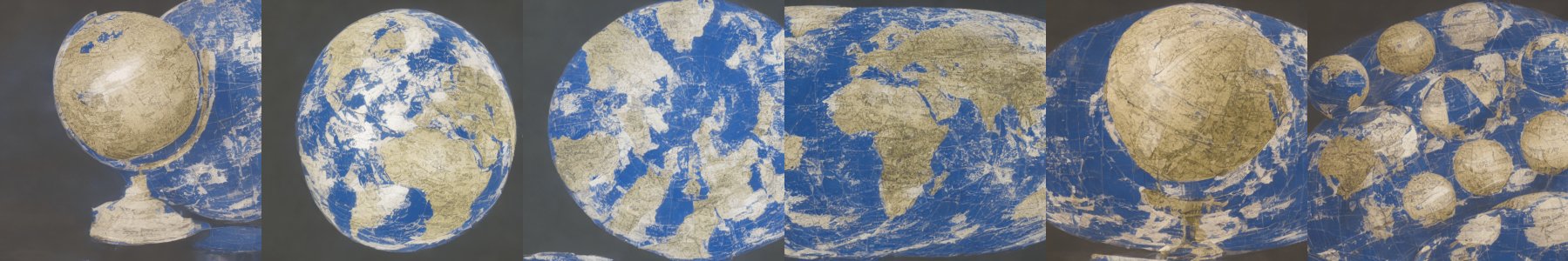}
    \end{minipage}

    \vspace{1em}

    \begin{minipage}[t]{\textwidth}
        \centering
        \textbf{FreeBloom} \\
        \includegraphics[width=0.85\linewidth]{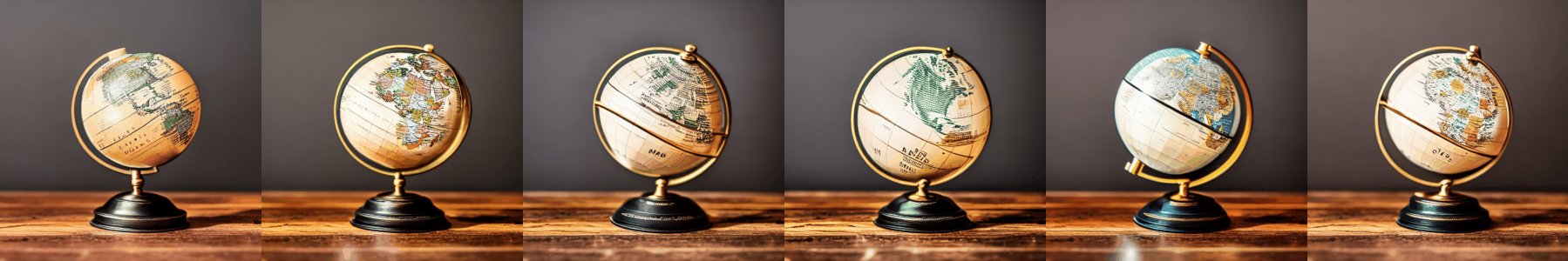}
    \end{minipage}

    \vspace{1em}

    \begin{minipage}[t]{\textwidth}
        \centering
        \textbf{EIDT-V SD} \\
        \includegraphics[width=0.85\linewidth]{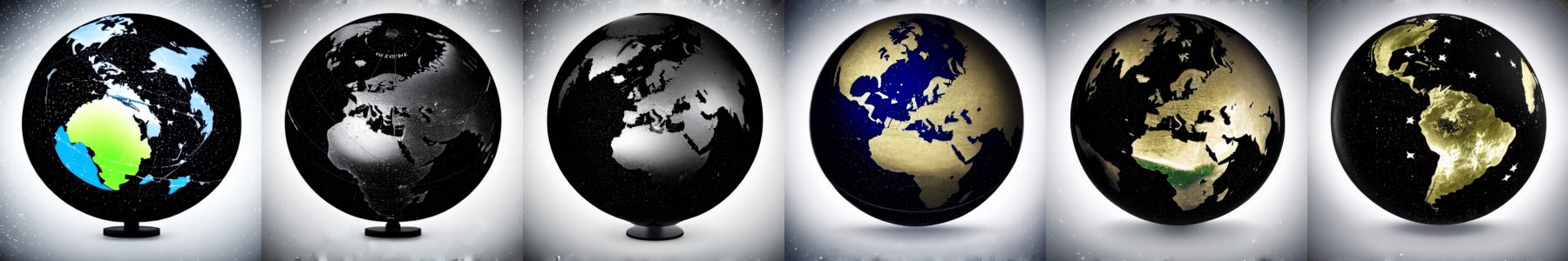}
    \end{minipage}

    \vspace{1em}

    \begin{minipage}[t]{\textwidth}
        \centering
        \textbf{EIDT-V SD\_IP} \\
        \includegraphics[width=0.85\linewidth]{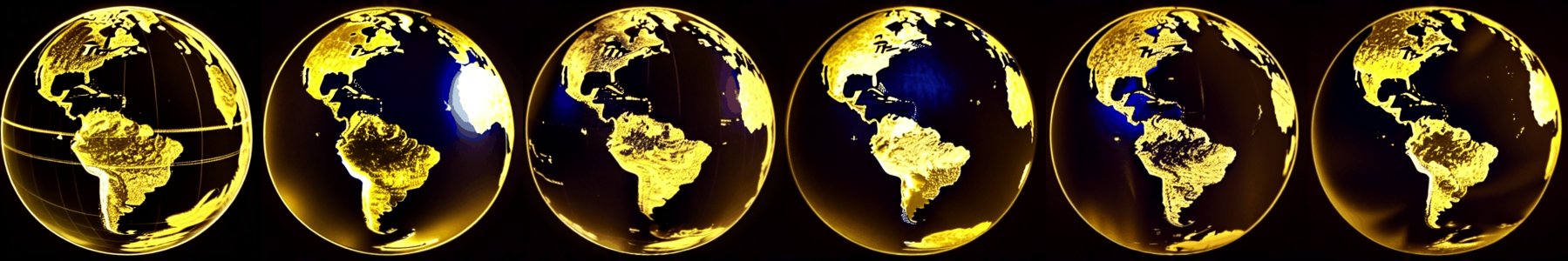}
    \end{minipage}

    \vspace{1em}

    \begin{minipage}[t]{\textwidth}
        \centering
        \textbf{EIDT-V SDXL} \\
        \includegraphics[width=0.85\linewidth]{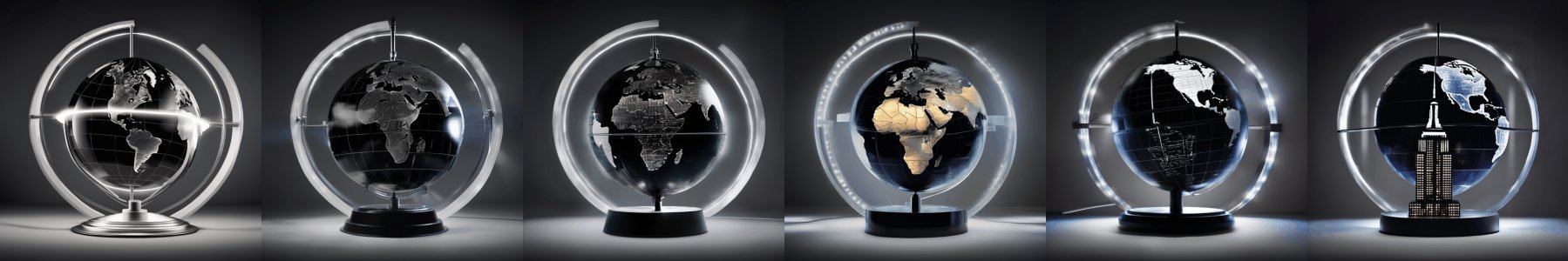}
    \end{minipage}

    \vspace{1em}

    \begin{minipage}[t]{\textwidth}
        \centering
        \textbf{EIDT-V SD3} \\
        \includegraphics[width=0.85\linewidth]{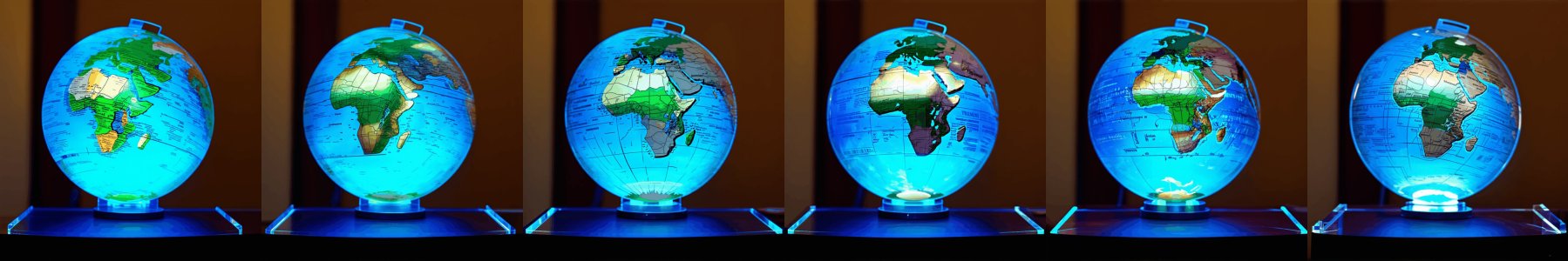}
    \end{minipage}

    \caption{A spinning globe showing continents passing by.}
\end{figure}

\begin{figure}[H]
    \centering
    \begin{minipage}[t]{\textwidth}
        \centering
        \textbf{T2VZero} \\
        \includegraphics[width=0.85\linewidth]{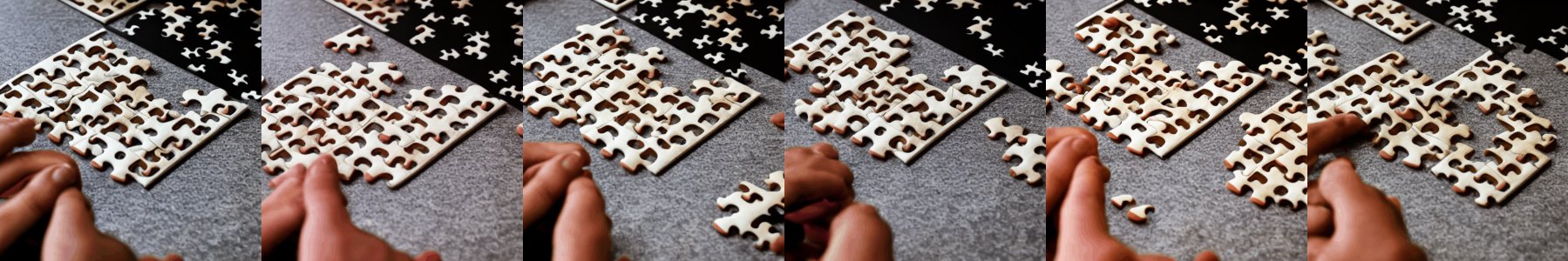}
    \end{minipage}

    \vspace{1em}

    \begin{minipage}[t]{\textwidth}
        \centering
        \textbf{DirecT2V} \\
        \includegraphics[width=0.85\linewidth]{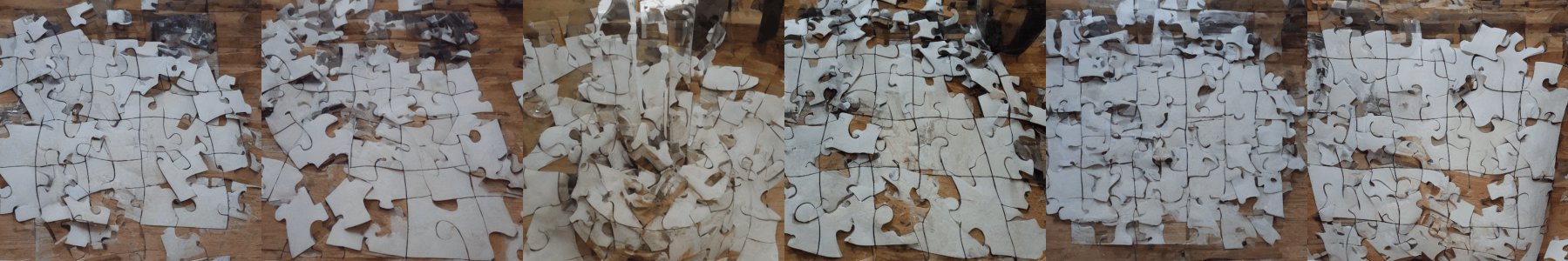}
    \end{minipage}

    \vspace{1em}

    \begin{minipage}[t]{\textwidth}
        \centering
        \textbf{FreeBloom} \\
        \includegraphics[width=0.85\linewidth]{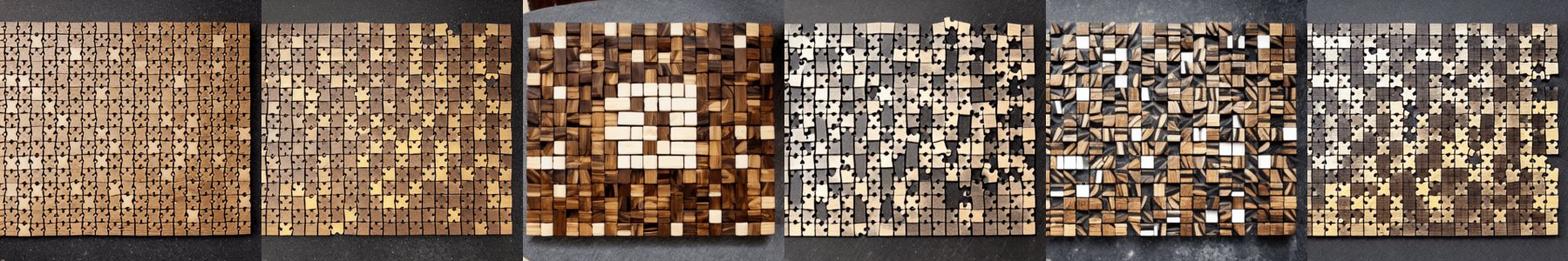}
    \end{minipage}

    \vspace{1em}

    \begin{minipage}[t]{\textwidth}
        \centering
        \textbf{EIDT-V SD} \\
        \includegraphics[width=0.85\linewidth]{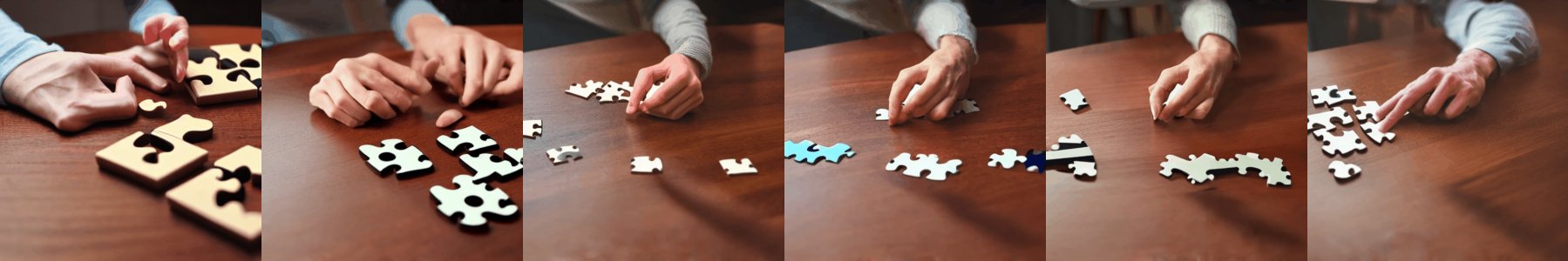}
    \end{minipage}

    \vspace{1em}

    \begin{minipage}[t]{\textwidth}
        \centering
        \textbf{EIDT-V SD\_IP} \\
        \includegraphics[width=0.85\linewidth]{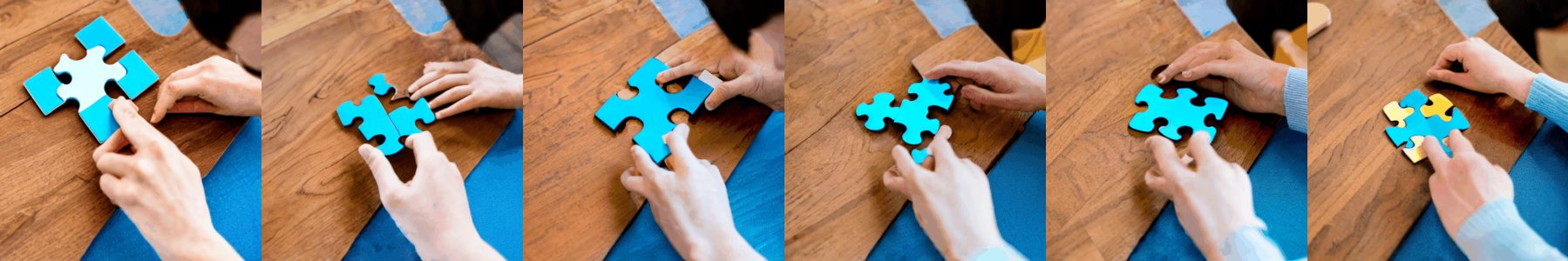}
    \end{minipage}

    \vspace{1em}

    \begin{minipage}[t]{\textwidth}
        \centering
        \textbf{EIDT-V SDXL} \\
        \includegraphics[width=0.85\linewidth]{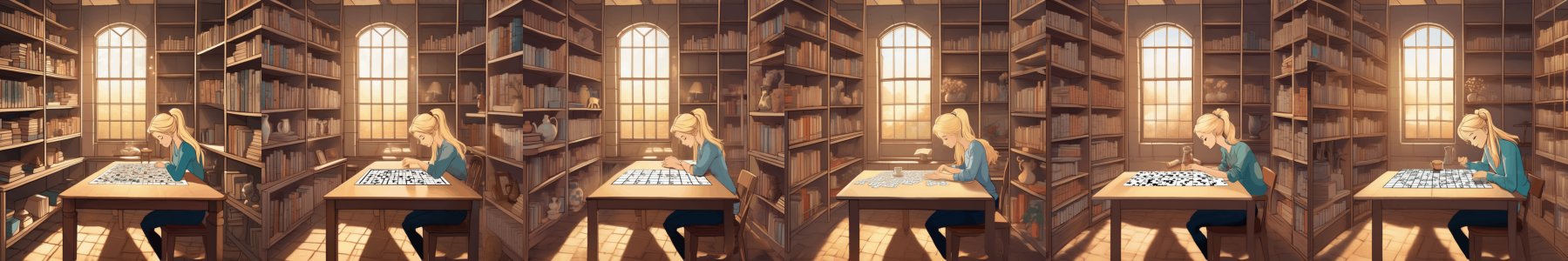}
    \end{minipage}

    \vspace{1em}

    \begin{minipage}[t]{\textwidth}
        \centering
        \textbf{EIDT-V SD3} \\
        \includegraphics[width=0.85\linewidth]{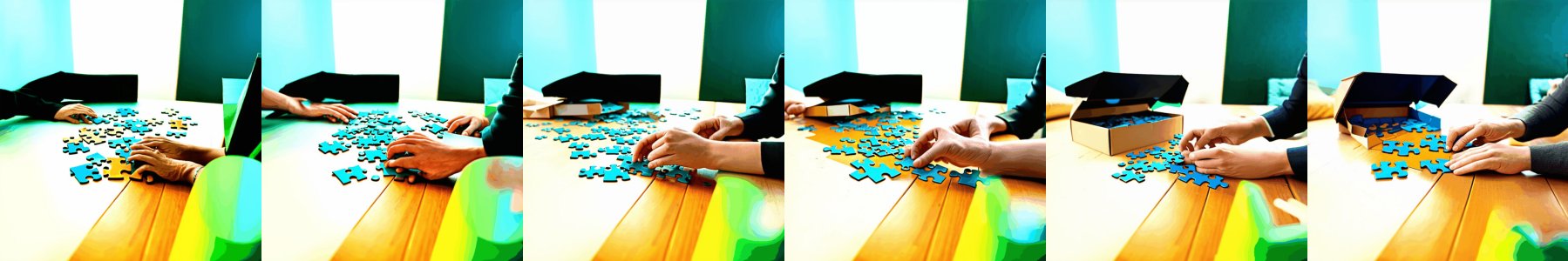}
    \end{minipage}

    \caption{A puzzle being assembled piece by piece.}
\end{figure}

%% file: Appendix/additional_best.tex
\section{Additional Best}
\label{sup:additional_best}

Here we highlight some of our best generations using the more powerful models SDXL and SD3. 
\subsection{SDXL}

\begin{figure}[H]
    \centering
    \includegraphics[width=\linewidth]{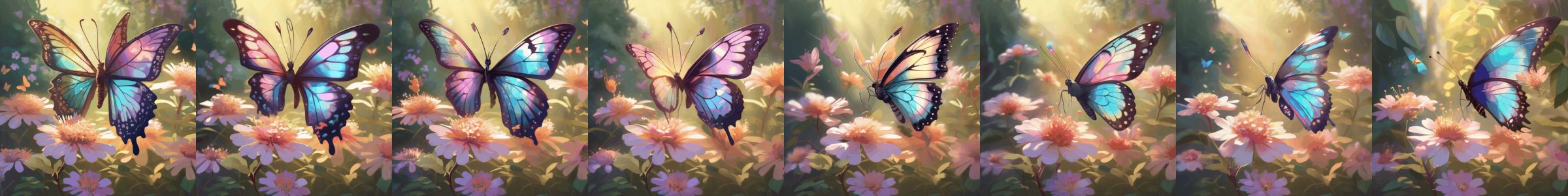}
    \caption{A butterfly gently flapping its wings while resting on a flower.}
\end{figure}

\begin{figure}[H]
    \centering
    \includegraphics[width=\linewidth]{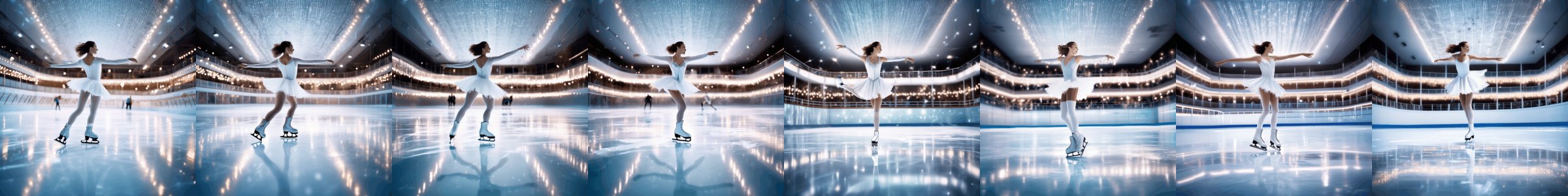}
    \caption{A figure skater gliding across an ice rink with smooth turns.}
\end{figure}

\begin{figure}[H]
    \centering
    \includegraphics[width=\linewidth]{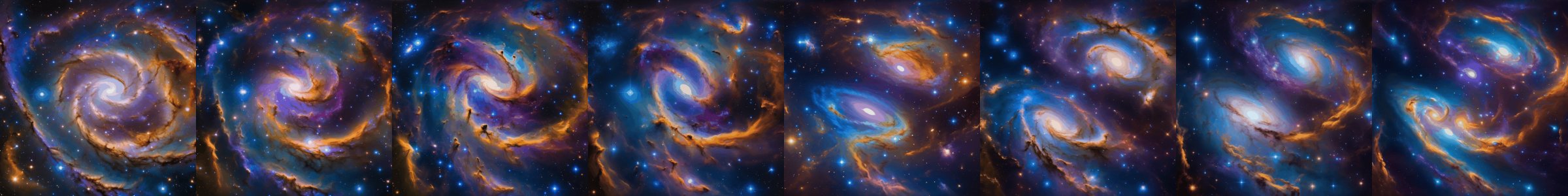}
    \caption{A galaxy swirling with stars and nebulae in deep space.}
\end{figure}

\begin{figure}[H]
    \centering
    \includegraphics[width=\linewidth]{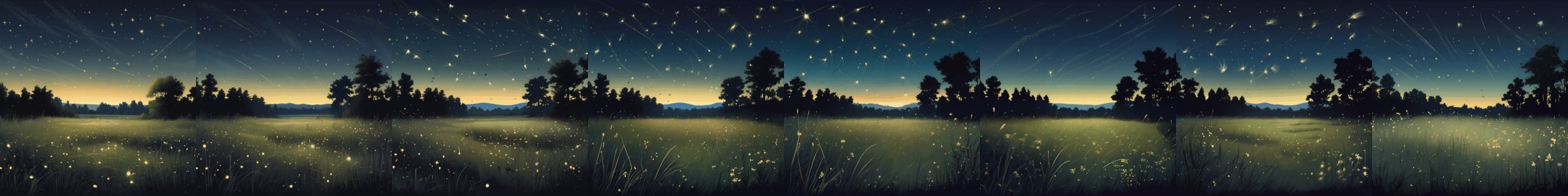}
    \caption{A lightning bug flying through a dark meadow.}
\end{figure}

\begin{figure}[H]
    \centering
    \includegraphics[width=\linewidth]{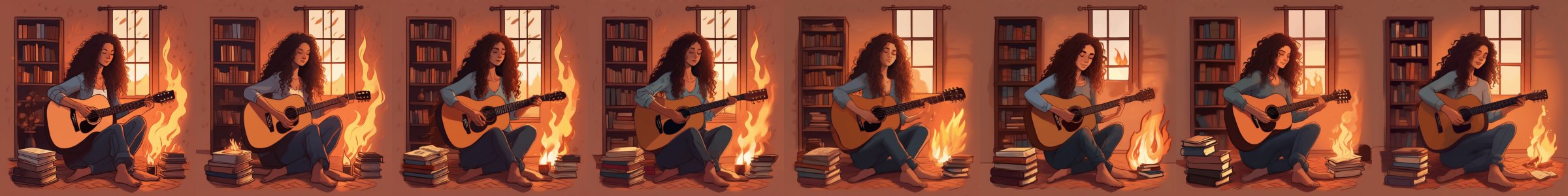}
    \caption{A musician playing a slow, peaceful tune on an acoustic guitar.}
\end{figure}

\begin{figure}[H]
    \centering
    \includegraphics[width=\linewidth]{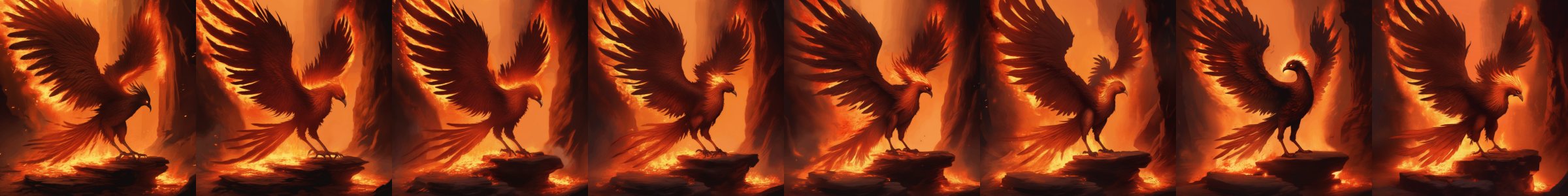}
    \caption{A phoenix slowly rising from glowing embers.}
\end{figure}

\begin{figure}[H]
    \centering
    \includegraphics[width=\linewidth]{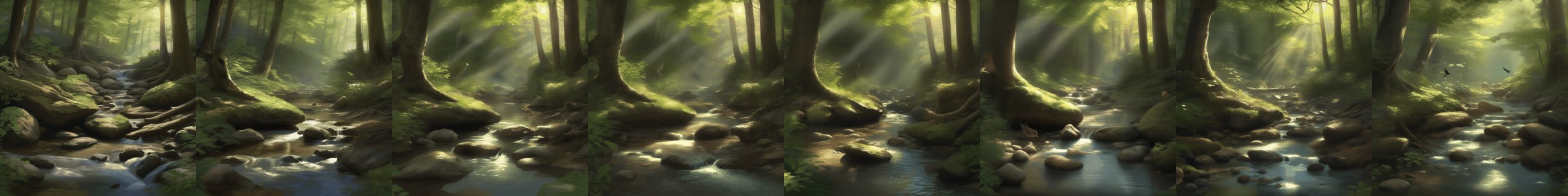}
    \caption{A stream flowing slowly over rocks in a forest.}
\end{figure}

\subsection{SD3 Medium}

\begin{figure}[H]
    \centering
    \includegraphics[width=\linewidth]{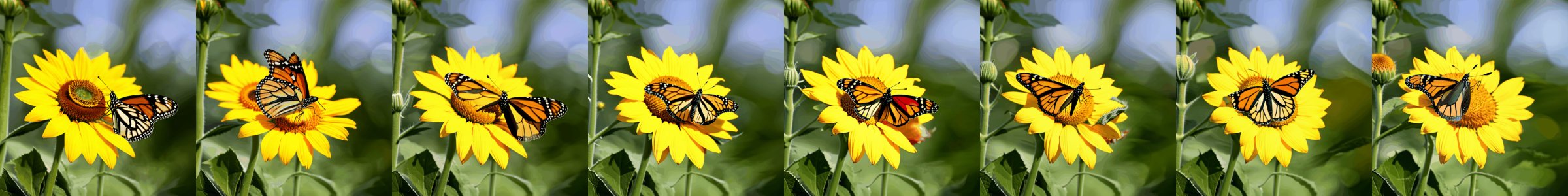}
    \caption{A butterfly gently flapping its wings while resting on a flower.}
\end{figure}

\begin{figure}[H]
    \centering
    \includegraphics[width=\linewidth]{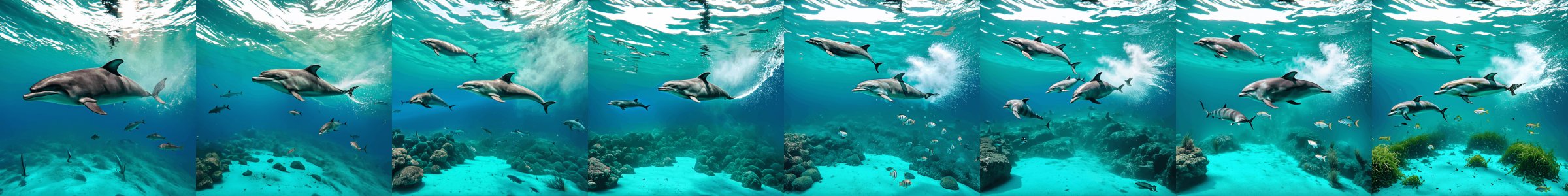}
    \caption{A dolphin gracefully gliding through turquoise waves.}
\end{figure}

\begin{figure}[H]
    \centering
    \includegraphics[width=\linewidth]{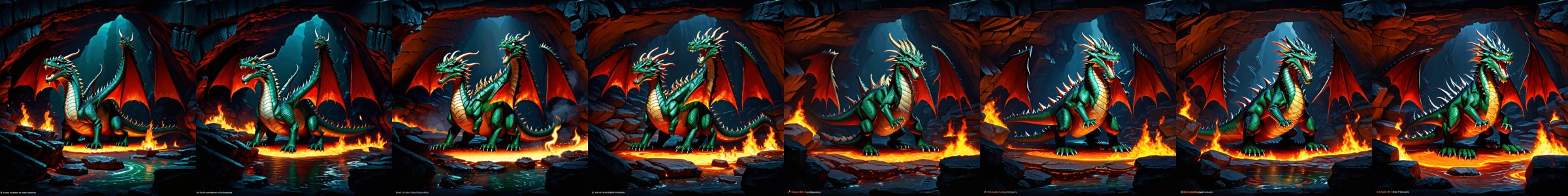}
    \caption{A dragon breathing a gentle stream of smoke from its nostrils.}
\end{figure}

\begin{figure}[H]
    \centering
    \includegraphics[width=\linewidth]{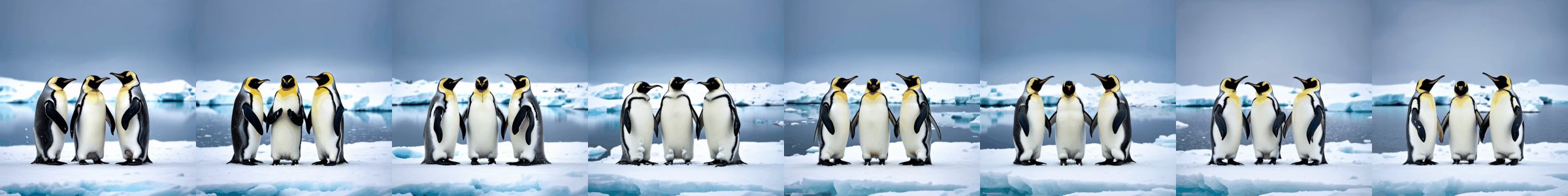}
    \caption{A family of penguins huddling together in a snowstorm.}
\end{figure}

\begin{figure}[H]
    \centering
    \includegraphics[width=\linewidth]{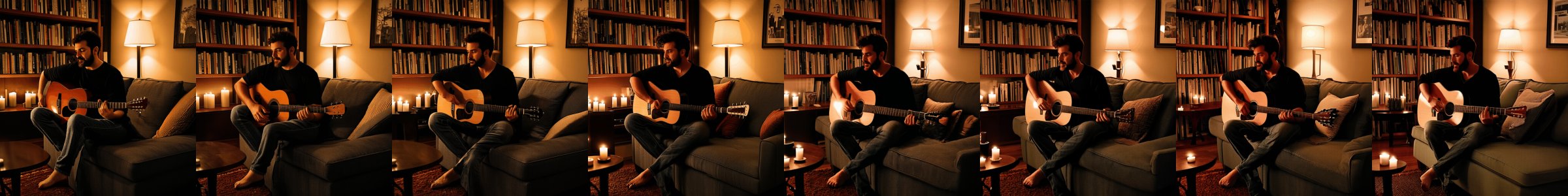}
    \caption{A musician playing a slow, peaceful tune on an acoustic guitar.}
\end{figure}

\begin{figure}[H]
    \centering
    \includegraphics[width=\linewidth]{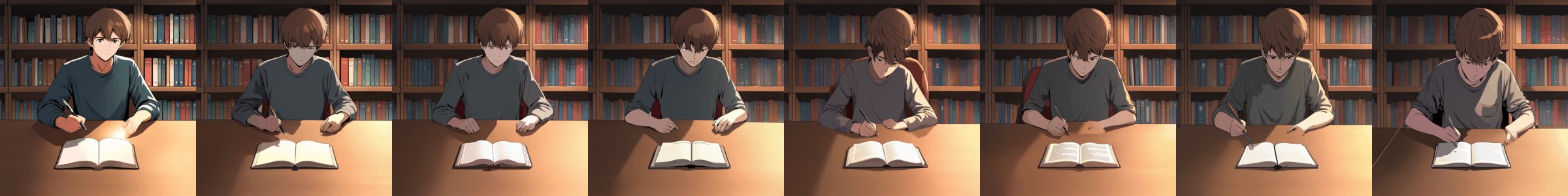}
    \caption{A person writing slowly in a journal with an ink pen.}
\end{figure}

\begin{figure}[H]
    \centering
    \includegraphics[width=\linewidth]{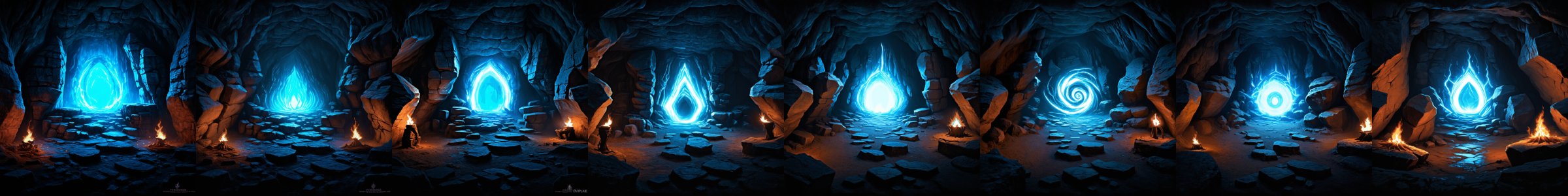}
    \caption{A portal opening and closing slowly in a mystical cave.}
\end{figure}

\begin{figure}[H]
    \centering
    \includegraphics[width=\linewidth]{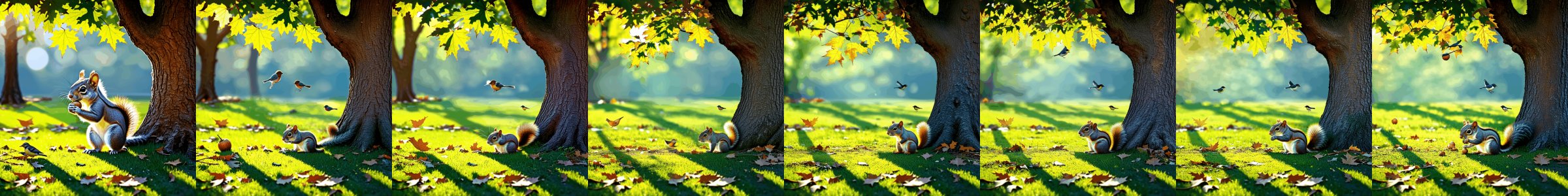}
    \caption{A squirrel nibbling on an acorn under a tree.}
\end{figure}

\begin{figure}[H]
    \centering
    \includegraphics[width=\linewidth]{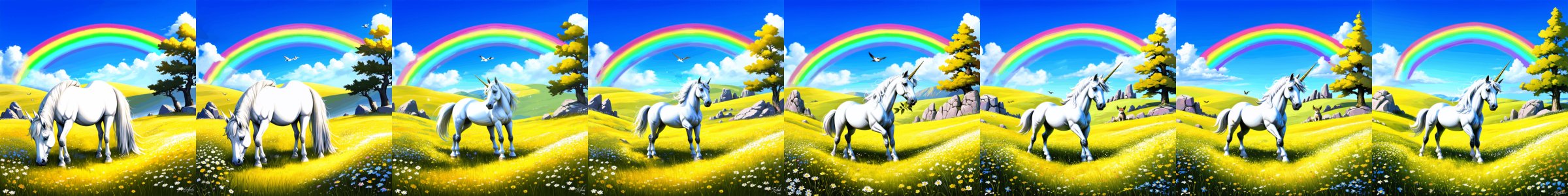}
    \caption{A unicorn grazing in a meadow under a rainbow.}
\end{figure}

\begin{figure}[H]
    \centering
    \includegraphics[width=\linewidth]{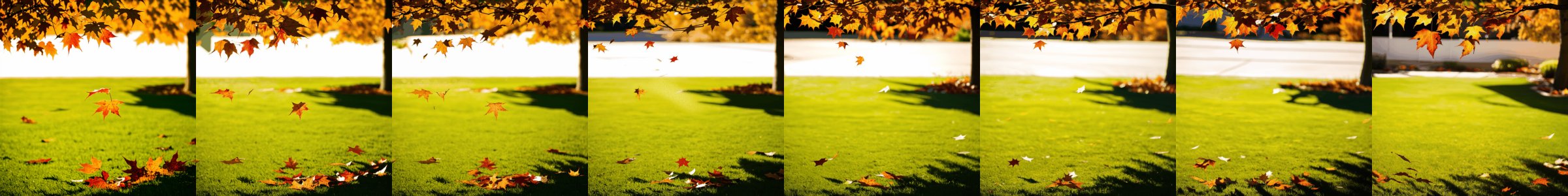}
    \caption{Golden leaves swirling softly in the autumn wind.}
\end{figure}